\documentclass[11pt]{article}

\usepackage[final]{acl}

\usepackage{times}
\usepackage{latexsym}

\usepackage[T1]{fontenc}

\usepackage[utf8]{inputenc}

\usepackage{newunicodechar}
\newunicodechar{》}{"{}"}

\usepackage{microtype}

\usepackage{inconsolata}

\usepackage{graphicx}

\usepackage{amsmath}
\usepackage{float}
\usepackage{fontawesome5}
\usepackage{multirow}
\usepackage{booktabs}
\usepackage{tabularx} 
\usepackage{array}    
\usepackage{ragged2e} 

\usepackage{amssymb}
\usepackage{pifont}
\usepackage{bbding}
\usepackage{wasysym}
\usepackage{enumitem}
\usepackage[dvipsnames,table,xcdraw]{xcolor}
\usepackage{subcaption}

\usepackage{tcolorbox}
\usepackage{makecell}
\usepackage{cleveref}
\usepackage{adjustbox}
\usepackage[ruled,linesnumbered]{algorithm2e}
\usepackage{algorithmic}

\crefformat{section}{#2§~#1#3}
\Crefformat{section}{#2§~#1#3}

\title{ProMediConv: Towards Proactive Conversational Mediation Agents for Multi-Party Dispute Resolution}
\title{ProMediConv: Benchmarking Proactive Conversational Agents in \\ Legal Dispute Mediation}

\author{
  Zesheng Wei$^1$\thanks{Work was done during a visit at SMU.} \quad
  Mengfan Li$^2$ \quad
  Wenhao Liu$^1$ \quad
  Yixin Zhang$^3$ \quad
  Zilei Wang$^1$\thanks{Corresponding author.} \quad
  Yang Deng$^2$ \\
  $^1$University of Science and Technology of China \quad
  $^2$Singapore Management University \\
  $^3$Institute of Artificial Intelligence, Hefei Comprehensive National Science Center \\
  \texttt{\{zswei, wenhaoliu\}@mail.ustc.edu.cn}, \texttt{mengfanli1024@gmail.com} \\
  \texttt{\{zhyx12, zlwang\}@ustc.edu.cn}, \texttt{ydeng@smu.edu.sg}
}

\begin{document}
\maketitle
\begin{abstract}
Dispute mediation is essential for maintaining social harmony and resilience, yet developing skilled mediators is costly and time-consuming. Existing LLM-based mediation research remains limited by unrealistic task formulations, low-fidelity datasets, and coarse evaluation metrics that obscure turn-by-turn dynamics. To address these gaps, we introduce ProMediConv, a novel benchmarking framework that models mediation as a proactive, multi-stage, and party-aware dialogue process incorporating 11 mediation strategies and four party behavior pattern (BP) states. Using 972 complete real-world cases, we construct a high-fidelity mediation dataset with utterance-level annotations of strategies and BP states. Furthermore, to better assess agent impact, we propose MAD (Mean Attribute Difference), a fine-grained metric that captures BP shifts throughout the dialogue. Leveraging this framework, we establish a comprehensive benchmark by evaluating diverse models alongside our tailored baseline ProMediAgent. Extensive empirical analyses reveal critical behavioral phenomena and underscore the persistent challenges current models face in dynamic, multi-party mediation. Ultimately, ProMediConv provides a rigorous foundation and a vital quantitative standard for advancing AI-assisted conflict resolution. Our dataset and codebase are accessible at \url{https://github.com/ZsWei66/ProMediConv_repo}.
\end{abstract}

\section{Introduction}
\label{sec:introduction}
Legal Dispute Mediation is a vital, non-litigious mechanism for resolving conflicts and maintaining social harmony \cite{mnookin1998alternative, di2009developing}. However, effective mediation demands far more than basic legal knowledge, it involves navigating complex, multi-party social interactions using advanced strategic communication and psychological assessment skills \cite{bercovitch1993evaluating}. Consequently, cultivating highly skilled human mediators is exceptionally resource-intensive and time-consuming \cite{devinatz2018makes}, creating a pressing need for automated solutions that can bolster societal capacity for conflict resolution.

\begin{figure}[t]
    \centering
    \includegraphics[width=\linewidth, trim=5pt 5pt 5pt 5pt, clip]{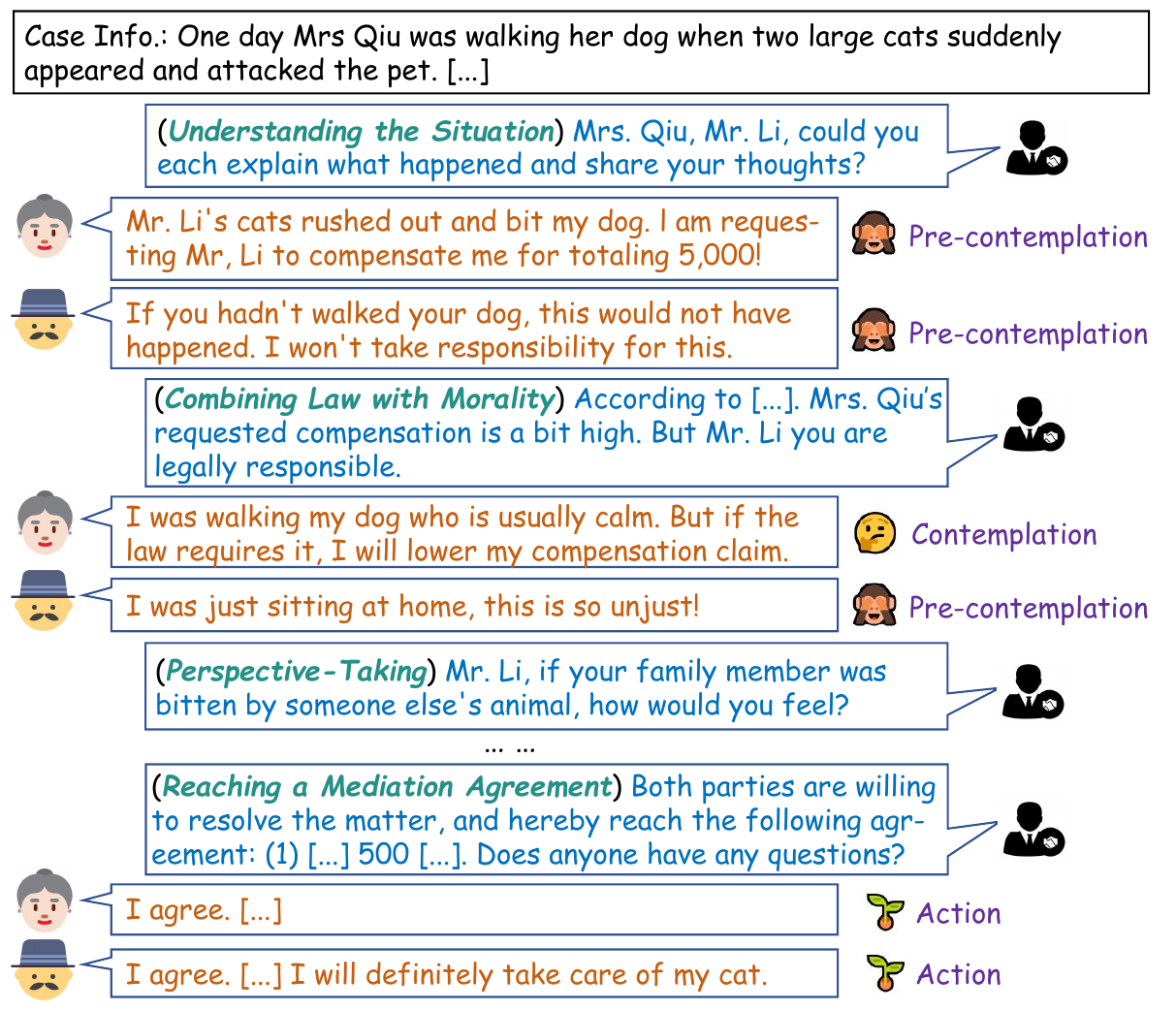} 
    \caption{An example dialogue of ProMediConv showcases the the mediator agent (\textcolor[rgb]{0, 0.439, 0.753}{Right}) proactively employ mediation strategies (\textbf{\textcolor[rgb]{0.141, 0.565, 0.529}{highlighted in brackets}}) to guide all disputing parties (\textcolor[rgb]{0.776, 0.373, 0.063}{Left}) toward a resolution. The \textcolor[rgb]{0.388,0.180,0.608}{\textbf{BP state}} of dispute party is explicitly annotated.}
    \label{fig:dia_exam_1}
    \vspace{-3mm}
\end{figure}

The rapid advancement of large language models (LLMs) has transformed general-purpose dialogue systems \cite{liu2024deepseek, ouyang2022training, qwen25}, demonstrating unprecedented conversational fluency and adaptability. Building on this progress, research efforts have expanded into domain-specific conversational applications, accelerating deployment in sectors such as persuasion \cite{wang2019persuasion}, legal consultation \cite{cui2023chatlaw}, and mental health support \cite{liu2021towards}. While recent studies have pioneered the use of LLM-based frameworks to explore automated mediation and negotiation \cite{chen-etal-2025-agentcourt}, fully expanding and quantifying the capabilities of conversational agents in this high-stakes domain requires a comprehensive benchmarking paradigm. Currently, systematic research in this direction remains severely constrained by several fundamental limitations.

\begin{figure*}[t]
\setlength{\abovecaptionskip}{5pt}   
\setlength{\belowcaptionskip}{0pt}
    \centering
    \includegraphics[width=\linewidth, trim=5pt 5pt 5pt 5pt, clip]{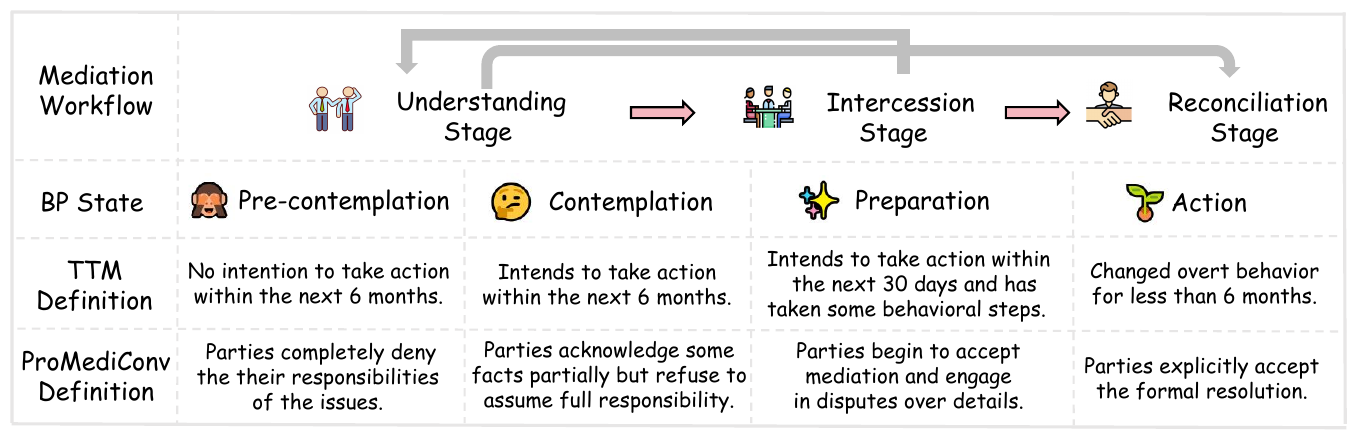} 
    \caption{Overview of the definitions of behavior pattern state of parties and mediation workflow in ProMediConv.
    \label{fig:bp_wor}
}
\vspace{-3mm}
\end{figure*}

Primarily, current frameworks exhibit a severe lack of alignment with real-world mediation dynamics. They often conceptualize mediation as a generic dialogue task \cite{bianchi2024llmsnegotiatenegotiationarenaplatform}, overlooking the mediator’s proactive, strategically directive role \cite{moore2014mediation, jiang2008mediation} and the complex multi-party communication inherent in the real-world scenarios \cite{10.1108/eb022898}. Compounding this issue is the scarcity of high-fidelity data. Due to privacy regulations, accessing authentic legal mediation transcripts is highly restricted. Consequently, existing studies are forced to rely on static legal documents lacking interactive dynamics \cite{DBLP:conf/naacl/HaleRCBG25}, or employ simplified LLM-based prompting pipelines to synthesize dialogues \citep{kwon-etal-2024-llms, zhang2024cpsycoun}. This results in small-scale, low-fidelity datasets that fundamentally fail to capture the nuanced, dynamic characteristics of authentic human mediation \cite{chen2025simulatingdisputemediationllmbased}.

Furthermore, current evaluation paradigms for proactive conversational agents \cite{deng2023survey} tend to be outcome-oriented. Existing metrics, such as Average Turn (AT), Success Rate (SR@$t$) and Soft Success Rate (SSR) \cite{deng2023plug, zhang-etal-2024-strength, DBLP:conf/aaai/HeL00SC0025, he2025simulating}, focus heavily on task efficiency and final resolution states. While useful for macroscopic assessment, they fail to capture within-dialogue qualitative impacts, such as the emotional improvement in emotion support, or the gradual shifts in dispute parties' psychological states during the mediation process. Capturing such user-side dynamics has increasingly attracted research attention \cite{DBLP:conf/acl/LiSD26, DBLP:conf/aaai/LiSD26}, as reflected in recent work on LLM-based user simulation \cite{DBLP:conf/acl/WuHGGYF26} or personality modeling \cite{DBLP:conf/acl/WeiLWD26, li2026llms} that influence users with different personality characteristics and behavioral states. Such process-level characterization is especially important in mediation, where relying solely on final outcomes can obscure how strategic interventions of mediators progressively shape the parties' psychological states and ultimately contribute to conflict resolution.

To this end, we systematically define \textbf{ProMediConv} (\underline{Pro}active \underline{Medi}ation \underline{Conv}ersation), a novel framework designed to benchmark proactive conversational agents in dynamic, strategy-driven mediation scenarios. We formally model the mediation workflow into three distinct stages, incorporating 11 widely adopted mediation strategies and four Behavior Pattern (BP) states to track the evolving psychological stances of the disputing parties. To support this task and overcome data scarcity, we construct a high-fidelity dataset from 972 real-world cases using an automated text reconstruction pipeline constrained by immutable factual and legal records. To resolve the evaluation bottleneck, we introduce MAD (Mean Attribute Difference), a fine-grained metric that quantifies mediation effectiveness by monitoring continuous changes in party BP.  Leveraging this framework, we establish a comprehensive benchmark by evaluating a diverse set of general and legal-specific LLMs with advanced policy planning methods, alongside our customized baseline ProMediAgent. Our extensive empirical analyses reveal critical behavioral phenomena and underscore the persistent challenges current models face in complex mediation environments. To summarize, our contributions are as follows:
\begin{itemize}[leftmargin=*]
\item We propose ProMediConv, systematically modeling multi-party dispute mediation dialogue task. We construct a high-fidelity dataset of 972 realistic cases, overcoming privacy barriers while preserving authentic interactive dynamics.

\item We introduce MAD, a novel metric designed to capture the within-dialogue qualitative shifts in parties' BP, addressing the blind spots of traditional efficiency- and outcome-focused metrics.

\item We benchmark diverse models alongside our tailored baseline, providing in-depth analysis of phenomenon in mediation dialogue and the inherent limitations of current conversational agents in real-world legal mediation.

\end{itemize}

\section{Related Works}
\textbf{Mediation and Negotiation Dialogues Frameworks}.~
Mediation and negotiation are common in social interaction, and have therefore attracted substantial attention in dialogue systems research. Prior work has largely focused on casual negotiation scenarios. For example, CraisglistBargain collected conversations involving prices bargaining of second‑hand items \cite{he2018decoupling}, LAMEN still target casual negotiations but broaden the range of tasks \cite{davidson2024evaluating}. More recently, researchers have begun to investigate mediation and negotiation dialogues in the legal domain, which represent more complex and professional scenarios \cite{chen-etal-2025-agentcourt}. However, existing frameworks suffer from overly simplified task formulations and weak alignment with real‑world dynamics, including restricting interactions to a two‑party setting \cite{hale-etal-2025-kodis}, failing to explicitly model the important party state \cite{robbennolt2025psychology, liu2025promediatesociocognitiveframeworkevaluating}, and neglecting the proactive, directive role that mediators often play in practice \cite{chen2025simulatingdisputemediationllmbased}. In addition, current mediation datasets are generally non‑scalable \cite{tan2024robots, chen2025simulatingdisputemediationllmbased, liu2025promediatesociocognitiveframeworkevaluating}, which constrains their usefulness for trainings and evaluations.

\noindent \textbf{Proactive Conversational Agents}.~
Most existing LLM‑based agents are designed to rigidly follow user instructions, which limits their ability to provide proactive assistance in certain scenarios \cite{10.1145/3539618.3594250, lu2024proactiveagentshiftingllm,DBLP:conf/acl/LinDWYD26}. To this end, proactive conversational agents are designed to anticipate potential impacts and actively steer the dialogue, rather than merely reacting to user inputs \cite{deng2023survey}.
Proactivity research in conversational agents typically focuses on two directions: 1) asking clarification question \cite{Zhu_2021,10.1145/3626772.3657869, DBLP:conf/sigir/ZhangWJSTC25}; 2) dialogue policy planning \cite{lei2022interactingnoncooperativeusernew, 10.1145/3626772.3657843}.
In the context of dialogue policy planning, existing research typically formulates proactivity as a strategy selection problem, where the agent chooses among conversational strategies at each turn \cite{dong-etal-2025-protod, DBLP:conf/aaai/HeL00SC0025}. For example, \cite{deng2023prompting} incorporate chain‑of‑thought into proactive dialogue, and  \cite{deng2023plug} introduce a planning plugin to enhance proactive capabilities. In addition, \cite{fu2023improving} improve strategic decision‑making through self‑play and learning from AI feedback. However, these approaches mainly focus on two-party conversational tasks, and, no prior work has explored evaluating proactive conversational agents in complex, multi‑party mediation scenarios.

\section{ProMediConv}
\label{sec:ProMediConv}
In this section, we formulate ProMediConv as a structured dialogue framework (with an example in Figure~\ref{fig:dia_exam_1} and complete formalization in \cref{app:ProMediConv_Task_Formalization}). This systematic formulation encompasses the mediation workflow and parties' BP states (\cref{sec:workflow_BP}), the mediation strategies (\cref{sec:strategy}), and our proposed MAD metric (\cref{sec:MAD}). Guided by this theoretical foundation, we subsequently detail the construction of the dataset (\cref{sec:dataset_construction}).

\subsection{Mediation Workflow \& BP States of Party}
\label{sec:workflow_BP}
We delineate the ProMediConv workflow into three primary stages: (I) Understanding, (II) Intercession, and (III) Reconciliation. While the standard progression is strictly sequential ((I)→(II)→(III)), the dynamic nature of mediation accommodates exceptional trajectories. For instance, mediators may need to revert to the Understanding stage for repeated reassurance ((II)→(I)), or bypass the Intercession stage for the straightforward resolution of minor disputes ((I)→(III)).

Crucially, driving this workflow requires tracking the internal states of the involved parties. Extensive research on dispute resolution highlights that the primary focus of human mediators lies in transforming the BP of dispute party \cite{Bush1994ThePO, boluwaduro2021mediation}. To formally model this, we incorporate a granular BP tracking mechanism into ProMediConv. Drawing inspiration from the Transtheoretical Model (TTM) \cite{Prochaska1997TheTM, prochaska1992attendance}, which posits that behavioral change is a gradual, multi-stage process, we categorize the parties' BP into four distinct developmental states. This classification aligns seamlessly with how parties' perspectives evolve under strategic mediation. Figure~\ref{fig:bp_wor} illustrates these BP states and their integration into the mediation workflow.

\subsection{Mediation Strategies}
\label{sec:strategy}
\begin{figure}[t]
\setlength{\abovecaptionskip}{5pt}   
\setlength{\belowcaptionskip}{0pt}
    \centering
    \includegraphics[width=\linewidth, trim=5pt 5pt 5pt 5pt, clip]{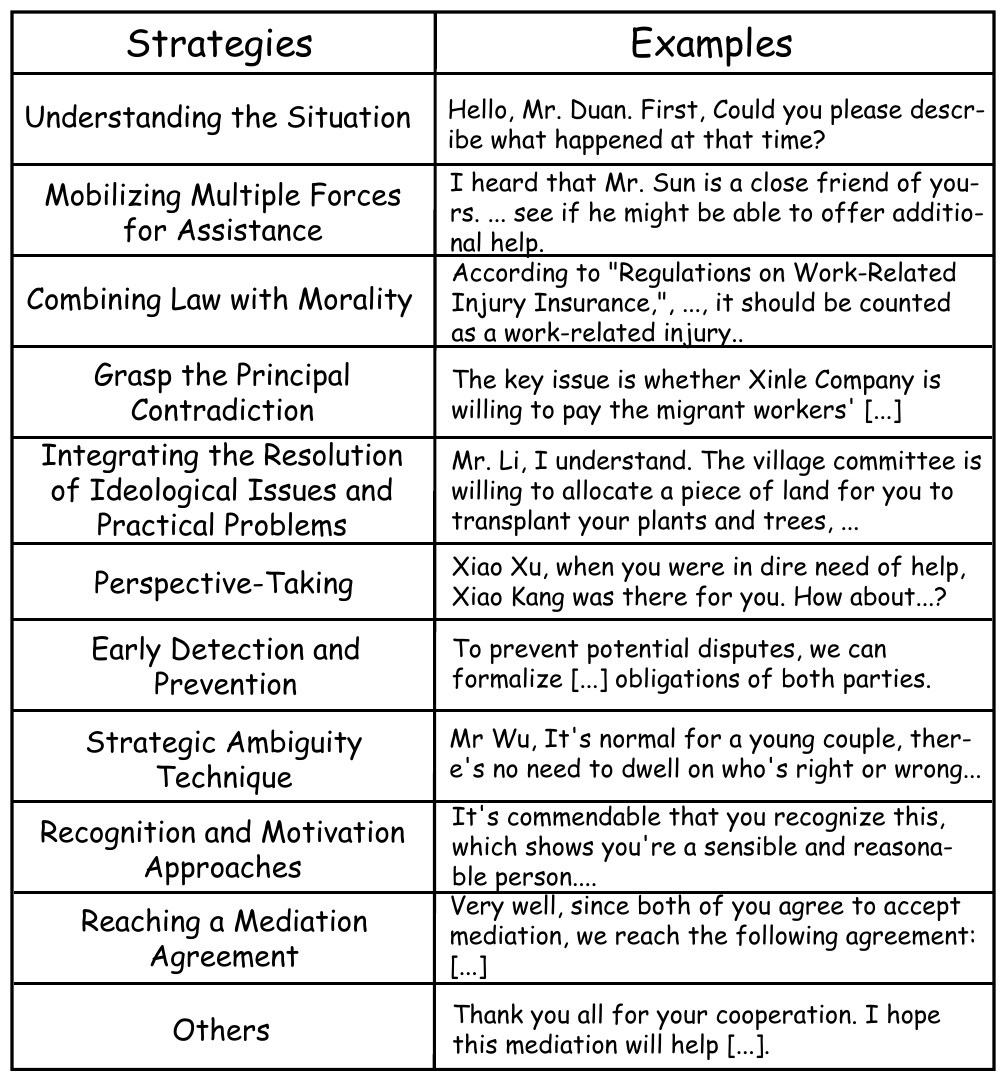} 
    \caption{Overview of Mediation Strategy Set. We abbreviate strategy names using first letter of their first two words (e.g., "Understanding the Situation" as "US"), except "Reaching a Mediation Agreement" as "RA".
    \label{fig:stra_set}
}
\vspace{-3mm}
\end{figure}

We synthesize a comprehensive set of mediation strategies grounded in widely used human mediation skills \cite{jiang2008mediation, writinggroup2017different, writinggroup2020methods}\footnote{These are official practitioner manuals published by the China Legal Publishing House. ISBNs and WorldCat links are provided in the references.}.
To ensure the completeness of the mediation process, we incorporate two additional strategies: ``Understanding the Situation'' and ``Reaching a Mediation Agreement''. 
Furthermore, to account for dialogues that do not explicitly employ specific mediation strategies, we categorize these under a strategy labeled ``Others''. 
The complete strategies set are provided in Figure~\ref{fig:stra_set}, with more detailed information provided in~\cref{sec:detailed_strategy_set}.

\subsection{Mean Attribute Difference}
\label{sec:MAD}

\begin{figure}[t]
\setlength{\abovecaptionskip}{5pt}   
\setlength{\belowcaptionskip}{0pt}
\centering
\includegraphics[width=\linewidth, trim=2pt 8pt 2pt 2pt, clip]{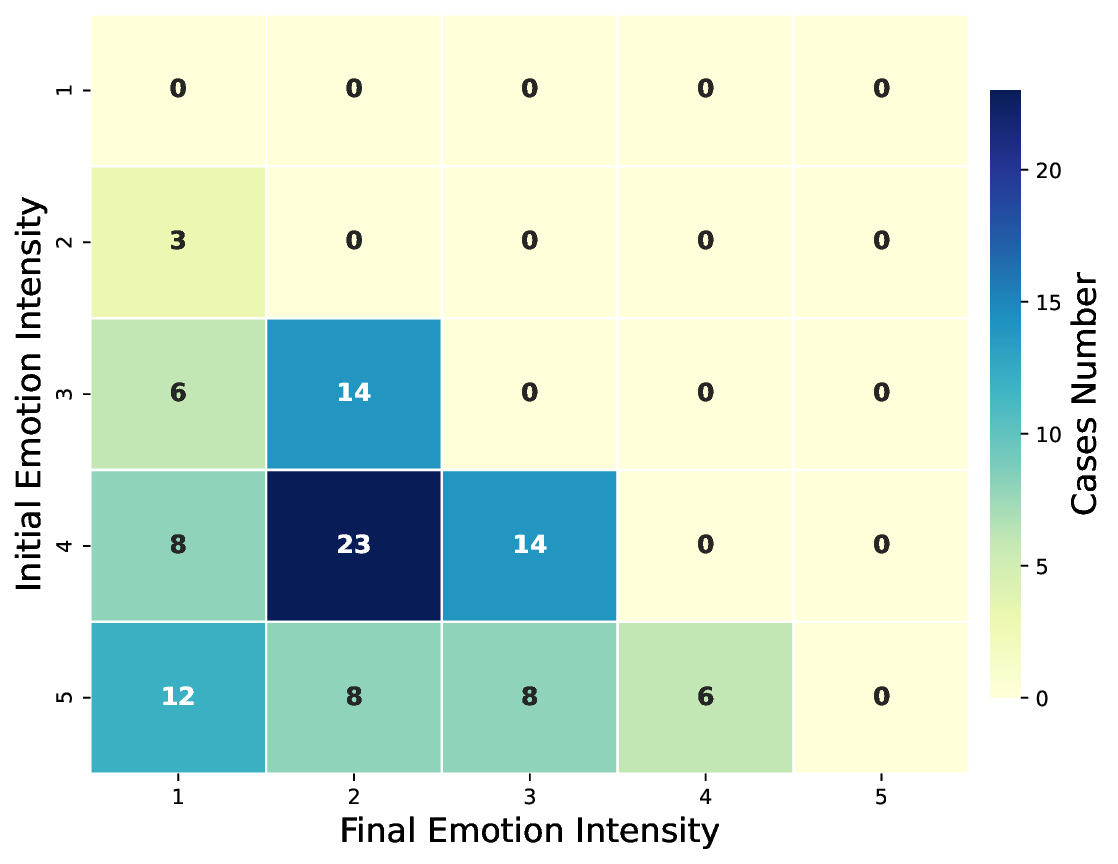}
\caption{Initial and final emotional intensity distribution (1–5) of help-seekers in successful dialogues (number of turns as 23) from the ESConv dataset.
\label{fig:emotion_heatmap}
}
\vspace{-3mm}
\end{figure}

As discussed in~\cref{sec:introduction}, traditional evaluation metrics for proactive dialogue agents predominantly focus on task efficiency and final outcomes, often obscuring fine-grained, within-dialogue qualitative impacts. To empirically substantiate this limitation, we analyze the ESConv dataset \cite{liu2021towards}. By isolating a subset of successful dialogues with an identical length (23 turns) to control for efficiency and outcome variables, we examine the distribution of the help-seekers' initial and final emotional intensities (rated on a 5-point scale). As illustrated in Figure~\ref{fig:emotion_heatmap}, even under identical macro-conditions (same turn count and successful resolution), there is substantial variance in the help-seekers' actual emotional intensity improvements. This explicitly demonstrates that dialogues appearing equally effective under traditional metrics can actually yield vastly different psychological benefits, thereby overlooking critical dimensions of an agent's true effectiveness.
To address this gap and comprehensively evaluate proactive conversational agents, we must consider not only task completion but also specific progressive improvements of the involved clients, such as help-seekers' emotional enhancement in psychology tasks, or the disputing parties' BP shifts in ProMediConv. Consequently, we propose \textbf{MAD} (\underline{M}ean \underline{A}ttribute \underline{D}ifference) as a more granular evaluation metric:
\begin{equation}
\label{eq:MAD_absolute}
\text{MAD} = \frac{1}{N} \sum_{k=1}^{N} \frac{\sum_{t=1}^{P_k} \left ( {a_t^f} - {a_t^i}\right )}{P_k}
\end{equation}
where \({a_t^i}\) and \({a_t^f}\) denote the initial and final states of client \(t\); \(P_k\) denotes the number of clients in the \(k\)-th case. Notably, to ensure the meaningfulness of the progress quantified by MAD, the evaluated client attribute must be formulated as ordinal, accommodating both discrete and continuous values.

\subsection{Dataset Construction}
\label{sec:dataset_construction}

Existing literature on dispute mediation preserves a wealth of authentic cases containing verbatim multi-party utterances and complete mediation records. However, utilizing these high-fidelity resources faces critical challenges: disorganized text formats, insufficient legal references, and a lack of explicit party state annotations. To construct the ProMediConv dataset without compromising the real-world dynamics, we design an automated text reconstruction pipeline (detailed in \cref{app:pipeline_details}) to standardize these cases into structured dialogues.

\subsubsection{Data Collection and Reconstruction}
We curate 981 real-world mediation cases from 15 published dispute mediation books via RapidOCR tools\footnote{\url{https://github.com/RapidAI/RapidOCR}}. After filtering out cases with low quality, we obtain 972 high-quality authentic cases. Instead of generating dialogues from scratch, our pipeline employs a strict constraint mechanism to prevent LLM hallucinations. Specifically, we first automatically extract the immutable facts, factual trajectories, and applicable legal clauses from the original unstructured texts. These extracted elements serve as hard constraints for the subsequent dialogue reconstruction, ensuring that the generated multi-party interactions strictly faithfully restore the original mediation dynamics. Furthermore, to automatically annotate the parties' BP states, we evaluate several advanced LLMs against expert human annotations (detailed in~\cref{app:bp_classifier}). We select Qwen2.5-14B-Instruct \cite{qwen25} as our BP annotator, as it demonstrated the highest alignment with human judgments. Since all source cases are published materials, the identities of all parties have been fully anonymized to protect privacy.

\begin{table}[h]
  \centering
  \small
  \begin{tabular*}{\columnwidth}{@{\extracolsep{\fill}}lccc}
    \toprule
    \textbf{Statistic}  & \textbf{Total} & \textbf{Mediator} & \textbf{Party}   \\
    \midrule
    Average Role Numbers  & 4.54   & 1     & 3.54 \\
    Number of Cases       & 972    & -     & -    \\
    Number of Utterances   & 17,277 & 8,774 & 8,503 \\
    \textit{Avg.} Utterances per Case  & 17.88  & 8.98  & 8.90 \\
    \bottomrule
  \end{tabular*}
  \caption{Statistical characteristics of ProMediConv.}
  \label{tab:dataset_statis}
  \vspace{-3mm}
\end{table}

\begin{figure*}[h]
    \centering
    \begin{subfigure}[b]{0.48\textwidth} 
        \centering
        \includegraphics[width=\linewidth, trim=2pt 2pt 2pt 2pt, clip]{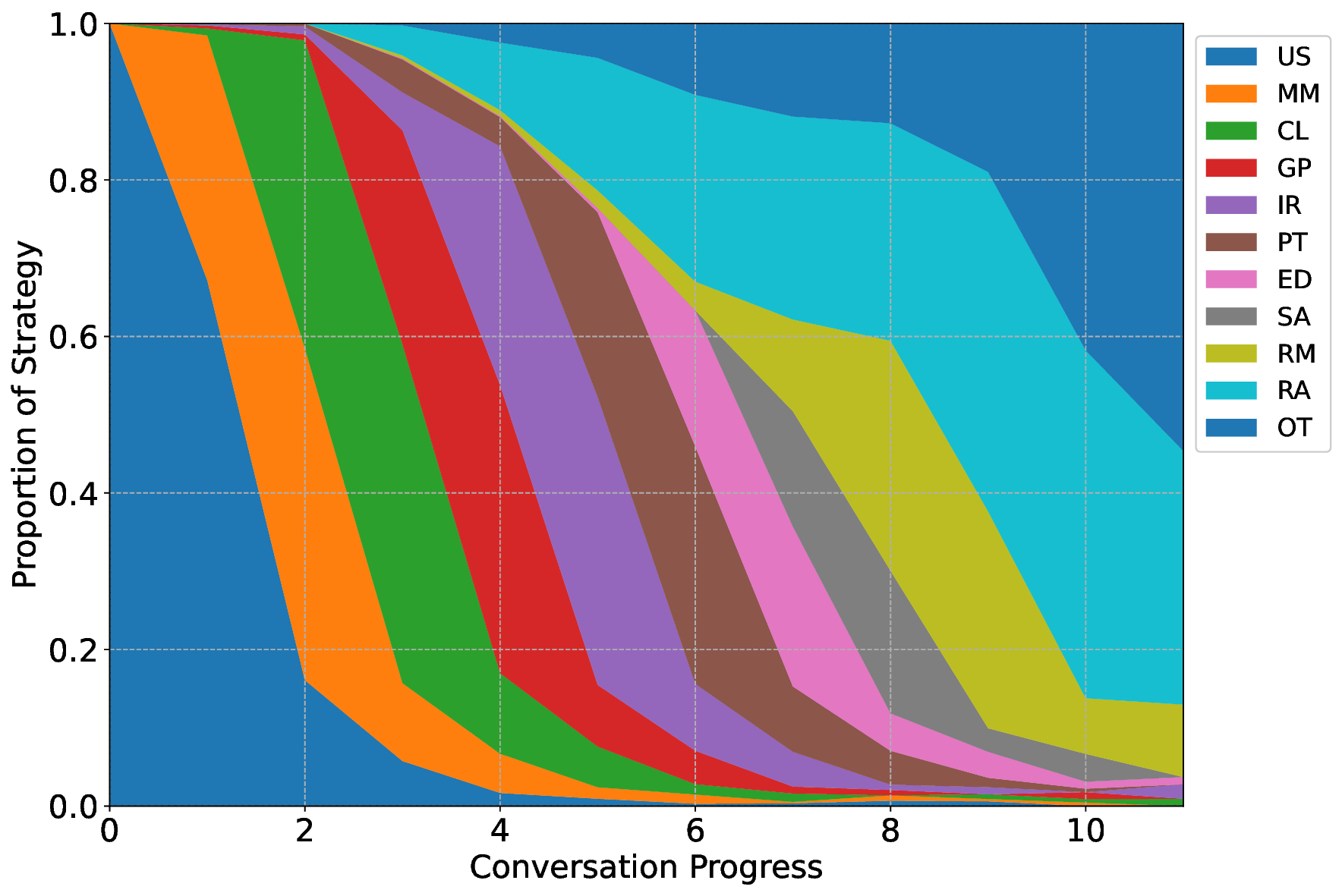} 
        \caption{}
        \label{fig:strate_dis}
    \end{subfigure}
    \hfill 
    \begin{subfigure}[b]{0.45\textwidth}
        \centering
        \includegraphics[width=\linewidth, trim=126pt 13pt 125pt 3pt, clip]{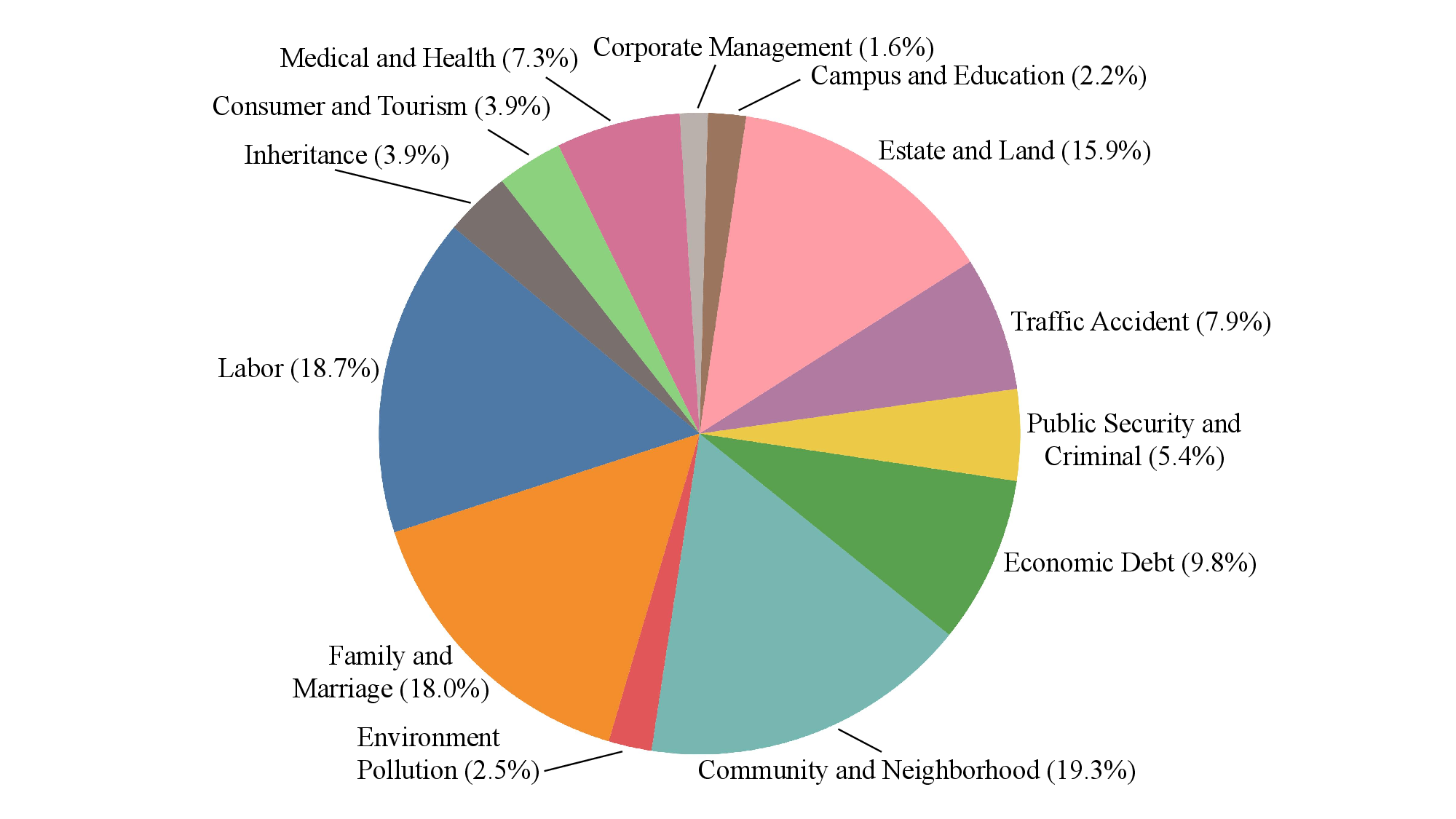}
        \caption{}
        \label{fig:dis_type}
    \end{subfigure}
    \vspace{-1mm}
    \caption{Dataset Statistics of ProMediConv. (a) Dynamic phase-based progression of mediation strategies across dialogue turns. (b) Long-tail distribution of the 13 dispute types, mirroring the real-world prevalence of civil cases.}
    \label{fig:data_statis}
\end{figure*}


\subsubsection{Dataset Statistics}
The final ProMediConv dataset comprises 972 structured multi-party dialogues with comprehensive BP and strategy annotations. As illustrated in Figure~\ref{fig:dis_type}, the dataset provides comprehensive coverage across 13 distinct dispute types. Rather than a strictly uniform distribution, it exhibits a realistic long-tail distribution that accurately reflects the real-world prevalence of civil disputes: primary categories such as Community and Neighborhood (19.3\%), Labor (18.7\%), Family and Marriage (18.0\%), and Estate and Land (15.9\%) constitute the majority, accompanied by a long tail of specific cases like Corporate Management (1.6\%).

Furthermore, Figure~\ref{fig:strate_dis} reveals the dynamic temporal evolution of mediation strategies across dialogue turns. This distribution highlights a clear phase-based progression, aligning seamlessly with our systematic modeling of strategy-stage correlations detailed in Table~\ref{tab:mediation_strategies_updated}. Rather than a static application, this dynamic evolution effectively captures the flexible, progressive, and highly contextual tactical choices made by professional mediators. Finally, we randomly partition the dataset into training, validation, and test sets at an 8:1:1 ratio for subsequent benchmarking.

\section{Overall Evaluations}
\label{sec:overall_evaluation}
In this section, we conduct comprehensive overall evaluations to benchmark proactive conversational agents on ProMediConv.

\subsection{Experimental Setups}

\subsubsection{Evaluation Metrics}
Following standard evaluation for proactive conversational agents, we employ AT, SR@$t$, and SSR:
\begin{equation}
\small
\label{eq:old}
\text{AT} = \frac{1}{N} \sum_{k=1}^{N} T_k, \quad \text{SR@}t = \frac{S}{N}, \quad \text{SSR} = \frac{1}{N} \sum_{k=1}^{N} r_k^f
\end{equation}
where $t$ represents the maximum dialogue turn limit; $N$ and $S$ denote the total number of cases and successfully resolved cases respectively; $T_k$ indicates the number of dialogue turns consumed in case $k$; and $r_k^f$ represents the reward at the final turn of case $k$.

Additionally, we adopt the MAD metric (Equation~\ref{eq:MAD_absolute}) to provide a more granular evaluation of the mediator agents' performance, specifically measuring their fine-grained influence on the disputing parties. For a clearer understanding of how each metric is calculated within ProMediConv, these formulations can be interpreted in conjunction with the fundamental workflow outlined in~\cref{app:ProMediConv_Task_Formalization}.

\subsubsection{Evaluated Models}
Given the absence of mediation-specific dialogue agents, we conduct a extensive evaluation using a diverse set of baselines.

\noindent \textbf{General LLMs} ~We evaluate Qwen2.5-7B-Instruct, Qwen2.5-14B-Instruct \cite{qwen25}, Llama-3.1-8B-Instruct \cite{dubey2024llama}, GLM-4-9B-0414 \cite{glm2024chatglm}, and ChatGPT (GPT-3.5-Turbo API) \cite{ouyang2022training}. To enhance their proactive mediation capabilities, we apply several advanced prompt-based policy planning methods, including Proactive \cite{deng2023prompting}, ProCoT \cite{deng2023prompting}, and ICL\_AIF \cite{fu2023improving}, alongside a vanilla standard setting.

\noindent \textbf{Legal-specific LLMs} ~To explore domain-specific performance, we select Wisdom-Interrogatory and Fuzi-Mingcha-v1.0 \cite{deng-etal-2023-syllogistic}, the top performers in the LawBench \cite{fei-etal-2024-lawbench} dialogue generation tasks (1-1, 3-5, 3-8). We evaluate them under standard prompts and Legal-enhanced Mediator Role-play (LMR) prompts.

\noindent \textbf{Our Trained Baseline} ~To explicitly validate the utility of the ProMediConv dataset, we introduce a customized baseline, named ProMediAgent. ProMediAgent employs a decoupled policy planner and response generator architecture \cite{he2018decoupling, deng2023plug}. We first conduct supervised fine-tuning (SFT) on the ProMediConv training set. Subsequently, we optimize the policy planner using the REINFORCE algorithm \cite{sutton1999policy} based on an AI-feedback scalar reward within a simulated mediation environment. The comprehensive details of the training objective and the reward mechanism are provided in \cref{app:promediagent}.

\begin{table*}[t]
  \centering
  \small
  \begin{tabular*}{\textwidth}{@{\extracolsep{\fill}}llccccr}  
    \toprule
    \textbf{Model} & \textbf{Method}  & \textbf{SR@\(t\)} \(\uparrow\) & \textbf{AT} \(\downarrow\) & \textbf{SSR} \(\uparrow\) & \textbf{MAD} \(\uparrow\)  & \textbf{\# of Tok.}  \\
    \midrule
    \multicolumn{7}{c}{\textbf{General LLMs}} \\ 
    \midrule
    \multirow{4}{*}{\textbf{Qwen2.5-7B-Instruct}} & Standard & 0.7532 & 12.43 & 0.8974 & 0.7576 & \(O(L)\)  \\
                               & Proactive & 0.7273 & 12.08 & 0.8452 & 0.7857 & \( O(2L) \) \\
                               & ProCoT & 0.6753 & 13.16 & 0.8039 & 0.8719 & \( O(2L) \) \\
                               & ICL\_AIF & 0.7670 & 12.06 & \underline{0.9154} & \textbf{1.0855} & \( O(3L) \) \\
    \midrule
    \multirow{4}{*}{\textbf{Qwen2.5-14B-Instruct}} & Standard & 0.7013 & 12.52 & 0.8056 & 0.9524 & \(O(L)\)  \\
                               & Proactive & 0.7922 & 11.56 & 0.9049 & 0.7446 & \( O(2L) \)  \\
                               & ProCoT & 0.7695 & 11.25 & 0.8733 & 0.9219 & \( O(2L) \)  \\
                               & ICL\_AIF & 0.7922 & 11.78 & 0.9071 & \underline{1.0443} & \( O(3L) \)  \\
    \midrule
    \multirow{4}{*}{\textbf{Llama-3.1-8B-Instruct}} & Standard & 0.7784 & 11.38 & 0.8469 & 0.7179 & \(O(L)\)  \\
                               & Proactive & 0.8096 & 11.09 & 0.9028 & 0.7926 & \( O(2L) \)  \\
                              & ProCoT & 0.8316 & 10.67 & 0.8674 & 0.8407 & \( O(2L) \)  \\
                              & ICL\_AIF & \underline{0.8589} & \underline{10.48} & 0.8597 & 0.8968 & \( O(3L) \)  \\
    \midrule
    \multirow{4}{*}{\textbf{GLM-4-9B-0414}} & Standard & 0.7167 & 12.41 & 0.8589 & 0.7722 & \(O(L)\)  \\
                               & Proactive & 0.8077 & 11.06 & 0.8654 & 0.7769 & \( O(2L) \)  \\
                              & ProCoT & 0.8359 & 10.87 & 0.9015 & 0.9115 & \( O(2L) \)  \\
                              & ICL\_AIF & 0.8179 & 11.27 & 0.8931 & 0.8675 & \( O(3L) \)  \\
    \midrule
    \multirow{4}{*}{\textbf{ChatGPT}} & Standard & 0.8517 & 10.67 & 0.9129 & 0.8202 & \(O(L)\) \\
                               & Proactive & 0.7953 & 11.36 & 0.8857 & 1.0074 & \( O(2L) \)  \\
                              & ProCoT & 0.8173 & 11.09 & 0.9094 & 0.9719 & \( O(2L) \)  \\
                               & ICL\_AIF & 0.8419 & 10.83 &0.9116  & 0.9481 & \( O(3L) \)  \\
    \midrule
    \multicolumn{7}{c}{\textbf{Legal-specific LLMs}} \\ 
    \midrule
    \multirow{2}{*}{\textbf{Fuzi-Mingcha-v1\_0}} & Standard & 0.6165 & 12.24 & 0.6386 & 0.4592 & \(O(L)\) \\
                                 & LMR & 0.6295 & 12.49 & 0.6827 & 0.4943 & \( O(2L) \)  \\
    \midrule
    \multirow{2}{*}{\textbf{Wisdom-Interrogatory}} & Standard & 0.5386 & 15.03 & 0.5128 & 0.7866 & \(O(L)\)  \\
                                 & LMR & 0.6646 & 13.06 & 0.6955 & 0.8434 & \( O(2L) \)  \\
    \midrule
    \multicolumn{7}{c}{\textbf{Our Trained Baselines}} \\ 
    \midrule
    \textbf{ProMediAgent}      & \multirow{3}{*}{\(-\)} & \textbf{0.9071} & \textbf{10.34} & \textbf{0.9428} & 0.9893 & \( O(L) \) \\
    ~ - w/o RL             & & 0.7692 & 11.64 & 0.8704 & 0.9673 & \( O(L) \)  \\
    ~ - w/o SFT+RL      &  & 0.6893 & 12.56 & 0.7715 & 0.8031 &\( O(L) \) \\
    \bottomrule
  \end{tabular*}
  
  \setlength{\abovecaptionskip}{5pt}   
  \setlength{\belowcaptionskip}{0pt}
  \caption{The overall evaluation results highlight that our dataset enables models with lower algorithmic complexity to perform better on ProMediConv. We bold the best results and underline the second-best ones in each column.}
  \label{tab:extrinsic_eval}
  \vspace{-3mm}
\end{table*}

\subsubsection{Implementation Details}
The interactive mediation environment for ProMediConv evaluation and the online learning of ProMediAgent adhere strictly to the workflow outlined in \cref{app:ProMediConv_Task_Formalization}. Following recent simulation works \cite{DBLP:journals/corr/abs-2603-03303}, we initialize simulated dispute parties using pre-annotated profiles and prompt the LLMs with tailored role-play templates to simulate dynamic interactions. An LLM-as-a-Judge mechanism determines the appropriate next speaker. Furthermore, an LLM-based outcome reward model evaluates the current mediation completion state. To mitigate generation stochasticity, we sample $l$ decoded sequences and compute the final scalar reward as their average \cite{DBLP:conf/iclr/0002WSLCNCZ23}. The maximum dialogue turn limit is fixed at 20. Regarding our proposed ProMediAgent, we employ Roberta-large\footnote{\url{https://huggingface.co/FacebookAI/xlm-roberta-large}} as the policy planner backbone and Qwen2.5-7B-Instruct as the response generator.  The SFT phase requires approximately 6 GPU hours over 45 epochs with a learning rate of $6 \times 10^{-6}$. The RL phase consumes roughly 16 GPU hours across 1,000 training episodes, employing a learning rate of $1 \times 10^{-6}$ and a discount factor $\gamma$ of 0.999. All training and inference procedures are conducted on a server equipped with eight NVIDIA L40S GPUs.

\subsection{Results and Analysis} 
Overall evaluation results are presented in Table~\ref{tab:extrinsic_eval}.

\paragraph{\textbf{Analysis of General LLMs}} As demonstrated, policy planning methods yield unstable effects on the proactive mediation capabilities of general LLMs. Certain methods even prove counterproductive across all ChatGPT settings. Furthermore, these minimal enhancements incur significant costs, as token consumption for mediator responses doubles with Proactive and ProCoT, and triples with ICL\_AIF. This indicates current methods may fail to fully capture essential mediation components or endow agents with stable proactive mediation capabilities. Consequently, these findings underscore a critical limitation within ProMediConv: even when leveraging large-parameter LLMs with sophisticated policy planning and increased token allocations, \textbf{the nuanced complexities inherent in ProMediConv remain elusive.}

\paragraph{\textbf{Analysis of Legal-Specific LLMs}} Both legal-specific LLMs perform poorly in standard settings, and even when augmented with applicable legal clauses reference, only minimal improvement is observed. Under standard conditions, the Wisdom model tends to produce generic responses that lead to conversational stagnation. We observe that both models configured with LMR successfully retrieve relevant legal provisions, which to some extent play positive impacts on involved parties. However, the improvement remains insufficient to achieve successful mediation. This phenomenon is reflected in the more significant improvements in SSR (↑ 6.91\%) and MAD (↑ 7.64\%) compared to SR@\(t\) (↑ 2.11\%), as exemplified by  Fuzi model. Moreover, in some cases, we observe that these models output irrelevant information due to overfitting. These findings highlight the limitations of current legal-specific models in ProMediConv and underscore the importance of our proposed MAD metric in detecting such subtle aspects of model performance.

\paragraph{\textbf{Analysis of ProMediAgent}} The ProMediAgent, trained via SFT on our ProMediConv dataset, exhibits significant improvements across all metrics (↑ 11.59\% SR@\(t\), ↑ 12.82\% SSR, ↓ 7.90\% AT, ↑ 0.44\% MAD). The performance of ProMediAgent is further optimized through RL, yielding gains of (↑ 17.93\% SR@\(t\), ↑ 8.32\% SSR, ↓ 11.17\% AT, ↑ 2.27\% MAD). This showcases notable advancements over the best-performing baseline, with SR@\(t\) increased by 5.61\%, SSR increased by 2.99\%, AT decrease by 1.34\%. ProMediAgent achieved MAD 8.86\% lower than the top-performing baseline. This margin is attributable to the reward function’s lack of explicit signals regarding changes in the party's BP. Those results validate that using our proposed ProMediConv enables models with fewer parameters and lower token usage to achieve performance superior with various LLMs with advanced policy planning methods.

\section{Further Analysis} 
In this section, we evaluate the impact of mediation strategies and present case studies on the Short-cut Resolution phenomenon to underscore MAD's necessity. Finally, we analyze model performance across varying party numbers and correlate MAD with current metrics.

\begin{table}[h]
\centering

\setlength{\abovecaptionskip}{5pt}
\setlength{\belowcaptionskip}{0pt}
\setlength{\tabcolsep}{4pt} 

\begin{adjustbox}{width=\linewidth}
  \begin{tabular}{llcccc} 
    \toprule
    \textbf{Model} & \textbf{Method} & \textbf{SR@\(t\)} \(\uparrow\) & \textbf{AT} \(\downarrow\) & \textbf{SSR} \(\uparrow\) & \textbf{MAD} \(\uparrow\) \\
    \midrule
    
    \multirow{2}{*}{\textbf{Qwen2.5-14B}} 
      & Qwen$_{\text{non}}$ & 0.7147 & 11.99 & 0.8479 & 0.6374 \\
      & Proactive     & 0.7922 & 11.56 & 0.9049 & 0.7446 \\
      
    \midrule
    
    \textbf{ProMediAgent} & - & \textbf{0.9071} & \textbf{10.34} & \textbf{0.9428} & \textbf{0.9893} \\
    \bottomrule
  \end{tabular}
\end{adjustbox}
\caption{Impact of mediation strategy constraints: Predefined taxonomy outperforms open-ended space.}
\label{tab:ablation_study}
\vspace{-3mm}
\end{table}

\subsection{Effect of Constrained Strategies Set}

We constrain the mediation agents to a taxonomy of 11 strategy categories rather than a fine-grained strategy space in ProMediConv. To validate the efficacy of such design, we conduct an empirical study using Qwen2.5-14B-Instruct with ablation setting \textbf{Qwen$_{non}$}, where the model generates unrestricted natural language strategy instruction, representing an open-ended strategy space. As illustrated in Table~\ref{tab:ablation_study},  Proactive method outperforms \textbf{Qwen$_{non}$}, indicating that a larger strategy search space does not necessarily yield improvements and may even degrade performance. This observation aligns with conclusions in relevant works~\cite{DBLP:conf/nips/Zeng0L24}, confirming that \textbf{our structured taxonomy serves as critical domain-specific priors to regularize the agent's action space}, preventing the instability inherent in unconstrained generation.

\begin{figure*}[t]
    \centering
    \begin{subfigure}[b]{0.49\textwidth} 
        \centering
        \includegraphics[width=\linewidth, trim=0pt 0pt 0pt 0pt, clip]{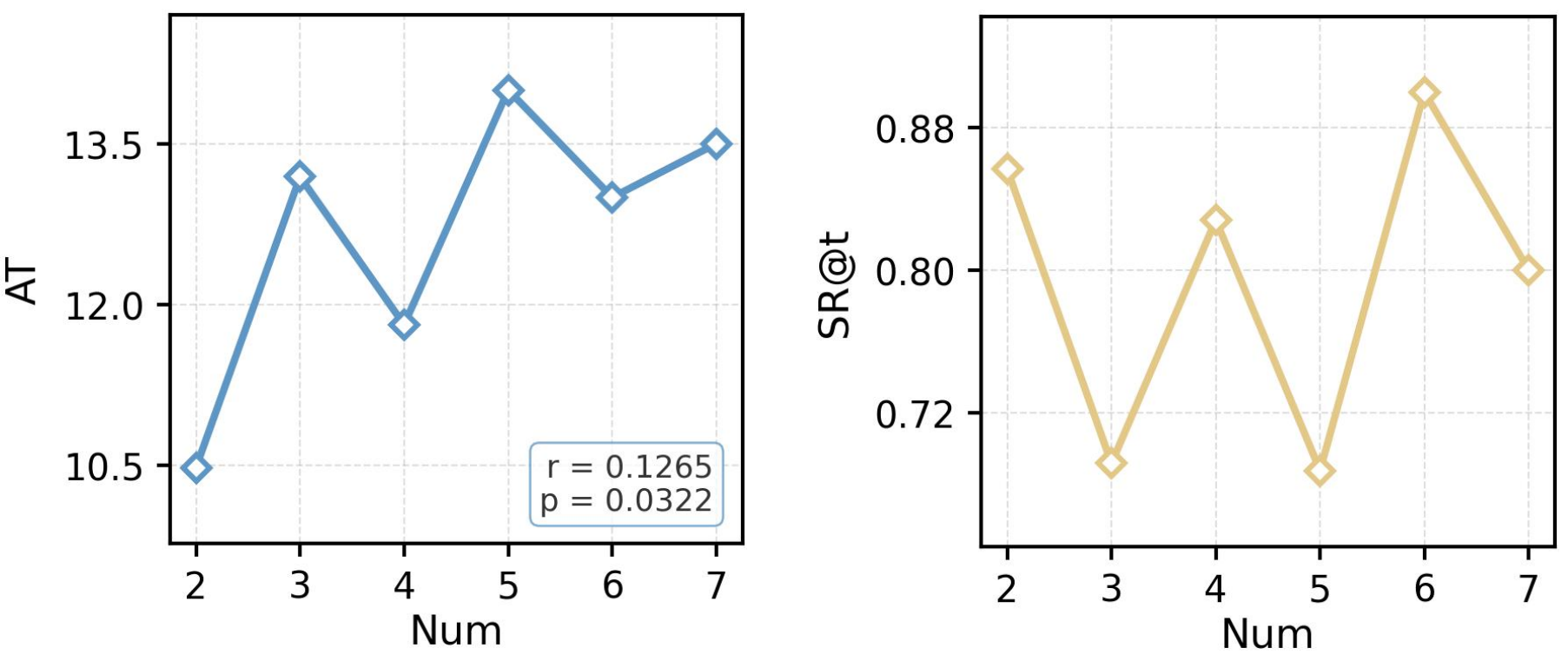}
    \end{subfigure}
    \hfill 
    \begin{subfigure}[b]{0.49\textwidth}
        \centering
        \includegraphics[width=\linewidth, trim=0pt 0pt 0pt 0pt, clip]{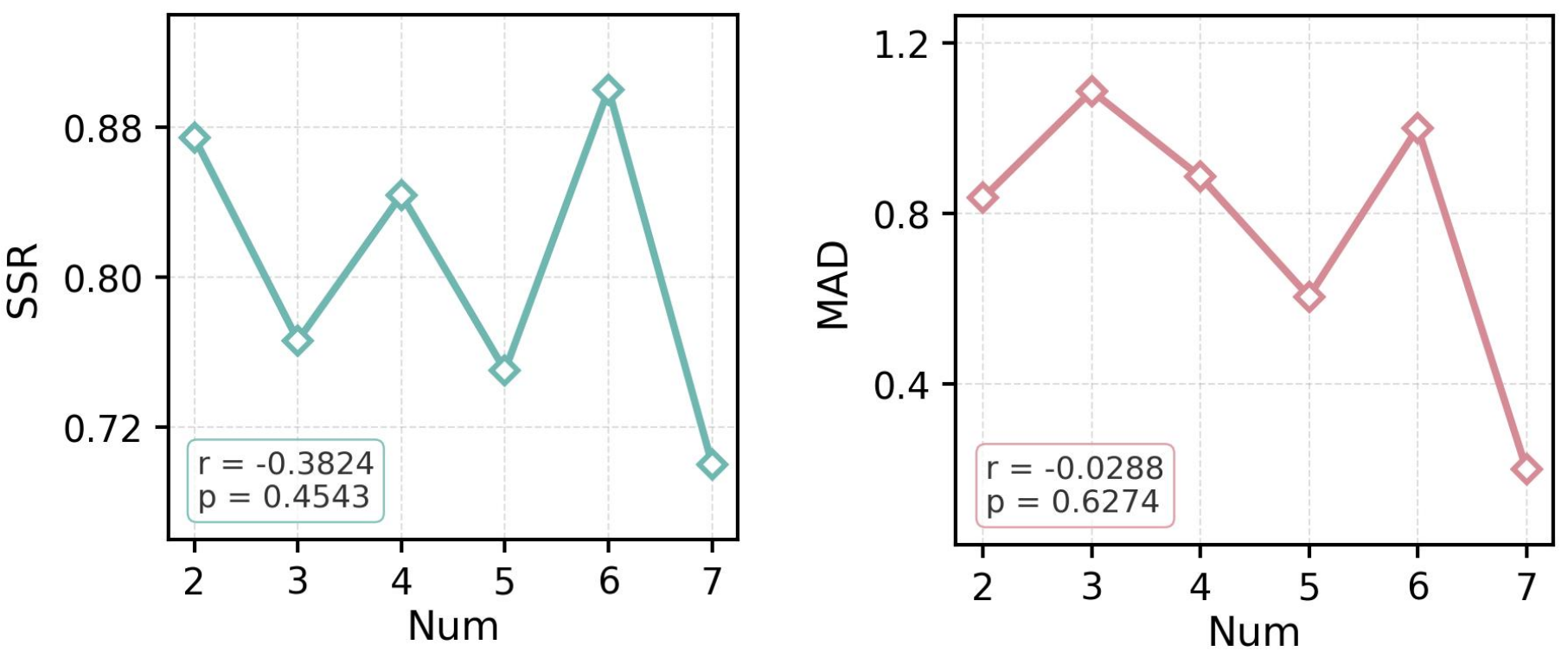}
    \end{subfigure}
    
    \vspace{-0.2cm}
    
    \caption{Correlation Analysis of Metrics and the Number of Parties. For metrics with continuous values (\textit{i.e.}, AT, SSR and MAD), Pearson coefficient and \(p\) value are reported. }
    \label{fig:metricVsPartyNum}
    \vspace{-3mm}
\end{figure*}

\subsection{Analysis on Mediation Strategy Impact} 
To investigate the impact of mediation strategies to dispute resolution, we evaluate ProMediAgent on our test set. Specifically, we establish two settings: (1) Random, where a strategy is selected randomly at each mediator turn. To ensure robustness, we report the average results of five independent rollouts per case; and (2) ProMediAgent, where strategies are determined by the policy planner of ProMediAgent, optimized via SFT and RL. Both settings employ the same response generator. We evaluate performance across all metrics and analyze variance across the five rollouts within the Random setting. Results are presented in Table~\ref{tab:rollout_analysis}. The Random baseline exhibits suboptimal performance due to stochastic selection, whereas ProMediAgent achieves significant improvements across all metrics through optimal strategy selection.  These findings validate that \textbf{optimal strategy selection is essential for enhancing mediation dialogue outcomes}, thereby substantiating the necessity of policy planning optimization to navigate the complex dynamics of ProMediConv.

\begin{table}[h]
\setlength{\abovecaptionskip}{5pt}    
\setlength{\belowcaptionskip}{0pt}
  \centering
    \small
  \begin{tabularx}{\linewidth}{@{} l *{4}{>{\centering\arraybackslash}X} @{}}
    \toprule
    \textbf{Metric} & \textbf{SR@\(t\)} \(\uparrow\) & \textbf{AT} \(\downarrow\) & \textbf{SSR} \(\uparrow\) & \textbf{MAD} \(\uparrow\) \\
    \midrule
    \textbf{Random} & 0.6769 & 12.64 & 0.8038 & 0.7692 \\
    \textbf{ProMediAgent} & 0.9071 & 10.34 & 0.9428 & 0.9893 \\
    \midrule
    \textbf{Gain (\(\Delta\))} & +34.0\% & +18.2\% & +17.3\% & +28.6\% \\
    \midrule
    
    \textbf{Variance} & 0.2581 & 5.2585 & 2.1989 & 0.6830 \\
    \bottomrule
  \end{tabularx}
  \caption{Impact of Mediation Strategy Selection. Significant gap between Random and Oracle highlights the impact of optimal mediation strategies in ProMediConv.}
  \label{tab:rollout_analysis}
  \vspace{-3mm}
\end{table}


\subsection{Case study on the necessity of MAD}
We identify a phenomenon, termed the Short-cut Resolution, in which mediator agents achieve superficial agreement without fully addressing all parties' claims. This issue arises as models exploit the reward model to prematurely conclude dialogues—often a result of inherent limitations in biased comprehension of LLMs. As exemplified in Table~\ref{tab:case_study}, ProMediAgent in Dialogue A takes more turns to reach a more thorough resolution, whereas in Dialogue B, the mediator uses fewer turns (better in AT metric), yet fails to fully address all claims, resulting in a lower MAD score. MAD effectively exposes such inadequacies among mediator agents by \textbf{granularly tracking BP states of each party, thereby validating its critical value}. 

\subsection{Performance w.r.t. the number of parties}
\label{sec:metric_vs_partiesnum}
Given the multi-party characteristic of ProMediConv, we further analyze how metrics vary with party numbers using all samples of Qwen2.5-14B-Instruct in ~\cref{sec:overall_evaluation}. As depicted in Figure~\ref{fig:metricVsPartyNum}, MAD and SSR show decreasing trends, while AT exhibits increasing trend as party numbers grow. This highlights that as case complexity increases, it becomes progressively harder for mediator agents to improve the situation and alter all parties' BP, thus necessitating more turns. Due to the large value of maximum  turns \(t=20\),  SR@\(t\) is not sensitive to the party numbers, exhibiting a horizontal pattern.


    

\subsection{MAD Correlation Analysis}
To validate the proposed MAD metric, we analyzed its correlation with existing metrics using the test samples consistent with \cref{sec:metric_vs_partiesnum}. Additionally, we recruit eight experienced human mediators to annotate completion state of those samples, establishing a Human Success Rate \textbf{(Human SR)}. Pairwise inter-annotator agreement reached 87\%, demonstrating a relatively high level of consistency. We computed Pearson and Spearman correlation coefficients between MAD and these metrics, as presented in Table~\ref{tab:mad_correlation}. The results indicate a weak correlation between MAD and AT. In contrast, MAD exhibits a significant positive correlation with SR@$t$. A notably stronger correlation is observed between MAD and SSR ($ p_r = 0.0022, p_\rho = 0.0046 $), likely because \textbf{both metrics facilitate fine-grained evaluation of the mediation process}. Critically, MAD demonstrates a stronger correlation with Human SR compared to SR@$t$ ($r=0.7386, \rho = 0.5838$). This alignment confirms that \textbf{higher MAD scores accurately reflect high-quality mediation from human perspective}, thereby substantiating the validity and robustness of our proposed MAD.

\begin{table}[h]
  \centering
  \small
  \begin{tabular}{lcccc}
    \toprule
    \makecell[l]{\textbf{Setting}} & \textbf{$r$} & $p_r$ & $\rho$  & $p_\rho$ \\
    \midrule
    \textbf{vs. AT}       & -0.2748 & 0.1654 & -0.2846 & 0.1502 \\
    \textbf{vs. SR@t}     &  0.6773 & 0.0057 &  0.4938 & 0.0421 \\
    \textbf{vs. SSR}      &  0.7847 & 0.0022 &  0.5281 & 0.0046 \\
    \midrule
    \textbf{vs. Human SR} &  0.7386 & 0.0039 &  0.5838 & 0.0548 \\
    \bottomrule
  \end{tabular}
  
  \caption{Statistical Analysis of MAD Correlations with other metrics. The Pearson and Spearman Correlations are denoted as $r$ and $\rho$, with statistical significance $p$.}
  \label{tab:mad_correlation}
  \vspace{-3mm}
\end{table}

\vspace{-1mm}

\section{Conclusion}

We introduce ProMediConv, a framework and high-fidelity dataset of 972 cases for benchmarking proactive, multi-party mediation. To address evaluation blind spots, we propose MAD, a metric quantifying effectiveness through shifts in user states. Finally, extensive benchmarking uncovers critical behavioral phenomena, exposing current LLMs' persistent limitations in complex real-world mediation.
Beyond methodological advancements, this study highlights the transformative potential of LLMs in fostering conflict resolution and social resilience. By establishing a scalable foundation for future AI-assisted mediation, our resources carry profound implications for enhancing community services, governance, and broader social harmony.

\section*{Limitations}
A primary limitation of ProMediConv is its exclusive reliance on text-based dialogue modeling within a specifically Chinese legal context. This inherently abstracts away non-verbal dynamics and may not fully capture the cultural nuances of dispute resolution in other languages. Future research should explore multi-modal integration and extend the framework to develop cross-cultural and cross-lingual mediation benchmarks. Nevertheless, this study serves as a vital foundational step. By delivering a scalable framework and rigorous benchmark, ProMediConv establishes a quantitative standard that paves the way for future AI-assisted mediation and broader social resilience.

\section*{Ethical Considerations}

This work utilizes open-source models and resources, including Llama-3.1-8B-Instruct, the Qwen2.5 series, GLM-4-9B-0414, RoBERTa-large, wisdomInterrogatory, and Fuzi-Mingcha-v1.0, strictly in accordance with their licenses and intended academic use. All mediation cases are sourced from published materials, with the identities of all involved parties fully anonymized to protect privacy. Human evaluators were explicitly informed of the data's purpose and consented to its exclusive use for academic research. ChatGPT is used only for limited paraphrasing and language polishing of author-written text.

\section*{Acknowledgments}

This research was supported by the National Natural Science Foundation of China under Grant 62406098 and the Anhui Province Key Research and Development Plan under Grant 202304a05020045. It is also supported by the National Research Foundation Singapore under the AI Singapore Programme (AISG Award No: AISG3-RPGV-2025-016). Yang Deng is supported by the Lee Kong Chian Fellowship awarded by Singapore Management University.

\bibliography{main}

\appendix




\section{Details of ProMediConv}

\subsection{Task Formalization}
\label{app:ProMediConv_Task_Formalization}

Based on all definitions in~\cref{sec:ProMediConv}, we model the mediation process in ProMediConv as a structured dialogue involving all parties and the mediator, with formalized illustrated in Table~\ref{alg:promediconv_flow}.
\begin{algorithm}[t]
\caption{Interaction Flow of ProMediConv}
\label{alg:promediconv_flow}
\begin{algorithmic}[1]
\REQUIRE Parties $P = \{p_1, \dots, p_n\}$, mediator \\$M$, strategies $S$, circumstances $R_i$
\ENSURE Total turns $T_k$, completion $S_k$, reward $r_{T_k}$, trajectories $\{b_t\}$, metrics
\STATE Init context $C_0 \leftarrow \text{Initial case description}$, turn $t \leftarrow 1$
\WHILE{Dialogue is not terminated}
    \STATE Speaker allocation: $s_t \leftarrow A(P \cup \{M\}, C_{t-1})$
    \IF{$s_t \in P$ \ (Dispute party selected)}
        \STATE Generate response via context: $u_t \leftarrow G_P(R_i, C_{t-1})$
        \STATE Identify behavior pattern: $b_t \leftarrow B(u_t, C_{t-1})$
        \STATE Update mediation context: $C_t \leftarrow C_{t-1} \oplus \{u_t\}$
    \ELSE[$s_t = M$ \ (Mediator selected)]
        \STATE Select mediation strategy: $\sigma_t \leftarrow \pi_M(S, C_{t-1})$
        \STATE Generate strategic utterance: $m_t \leftarrow G_M(\sigma_t, C_{t-1})$
        \STATE Update mediation context: $C_t \leftarrow C_{t-1} \oplus \{m_t\}$
        \STATE Evaluate state \& calc reward: $r_t \leftarrow R(C_t)$
    \ENDIF
    \STATE $t \leftarrow t + 1$
\ENDWHILE
\STATE $T_k \leftarrow t - 1$
\STATE Record completion $S_k \in \{0, 1\}$ and final reward $r_{T_k}$
\STATE Compute metrics (AT, SR@$t$, SSR, MAD) \\ via $T_k, S_k, r_{T_k}, \{b_t\}$
\RETURN $T_k, S_k, r_{T_k}, \{b_t\}$, alongside evalu-
ation metrics
\end{algorithmic}
\end{algorithm}

For each round \(t\), a party or the mediator will be invited to speak:
\begin{equation}
  \label{eq:parties_choose}
  p_i^t = A(P, C_{t-1}) \quad p_i^t \in P
\end{equation}
where \( P = \{p_1, p_2, \dots, p_n, M\} \)
represents the set of all involved parties and the mediator M, and \( C_{t-1} \) references the current mediation state. If it is the turn of a party to speak, he/she will speak based on their own circumstances \( R_{t} \) and in consideration of the current mediation state, and their behavior pattern \( b_i^t \)  at turn \( {t} \) will be marked by automated BP classifier agent \( B \):
\begin{equation}
  u_i^t = p_i^t(R_i, C_{t-1}), \quad
  b_i^t = B(u_i^t, C_{t-1})
\end{equation}

For mediator's turn, the mediator \( M \)  will first select a proper mediation strategy and then speak based on  the chosen strategy:
\begin{equation}
  \label{eq:strat_choose}
  \sigma_t = M(S, C_{t-1}) \quad \sigma_t \in S
\end{equation}
\begin{equation}
  \label{eq:mediator_speak}
  m_t = M(\sigma_t, C_{t-1})
\end{equation}
where \( S \) represents the mediation strategies set. And the mediation state C is updated with each successive utterance:
\begin{equation}
  \label{eq:history_update}
  C_t = C_{t-1} \sqcup \{u_i^t, m_t\}
\end{equation}

A reward model \(R\) will measure the current state and output a task-oriented reward to determine the completion state of the mediation goal, following \cite{deng2023plug}:
\begin{equation}
  \label{eq:reward_1}
  r^t = R(C_t )
\end{equation}
When the dialogue reaches termination, the total  turns \(T_k\), the completion state \(S_k \in \{0, 1\}\) (where \(S_k = 1\) denotes a success of the mediation goal and \(S_k = 0\) denotes a failure owing to the maximum turn limit) , the final-turn reward \(r_k^f\)  and the initial and final behavior patterns of all the parties \(\left\{ b_i^t \,\middle|\, i = \{1, 2, \ldots, n\},\ t = \{1,  T_k\} \right\} \) of current case \(k\)  will be recorded for the computation of dataset-level metrics AT, SR@\(t\), SSR and MAD.


\subsection{Complete Mediation Strategy Set in ProMediConv}
\label{sec:detailed_strategy_set}
As introduced in~\cref{sec:strategy}, we formalize a comprehensive taxonomy of 11 mediation strategies to capture the proactive and directive nature of human mediators. We presents the complete specification of this mediation strategy set in Table~\ref{tab:mediation_strategies_updated}. 

Specifically, for each mediation strategy, the table delineates its applicable mediation stages (I: Understanding, II: Intercession, and III: Reconciliation), provides an illustrative utterance example from authentic cases, and offers a detailed theoretical explanation of its core objective. This systematic mapping ensures that the defined strategies not only cover the entire lifecycle of a mediation session but also strictly align with the phase-based progression of real-world dispute resolution. 
\begin{figure}[t]
\setlength{\abovecaptionskip}{0pt}   
\setlength{\belowcaptionskip}{0pt}
    \centering
    \includegraphics[width=\linewidth, trim=5pt 2pt 5pt 5pt, clip]{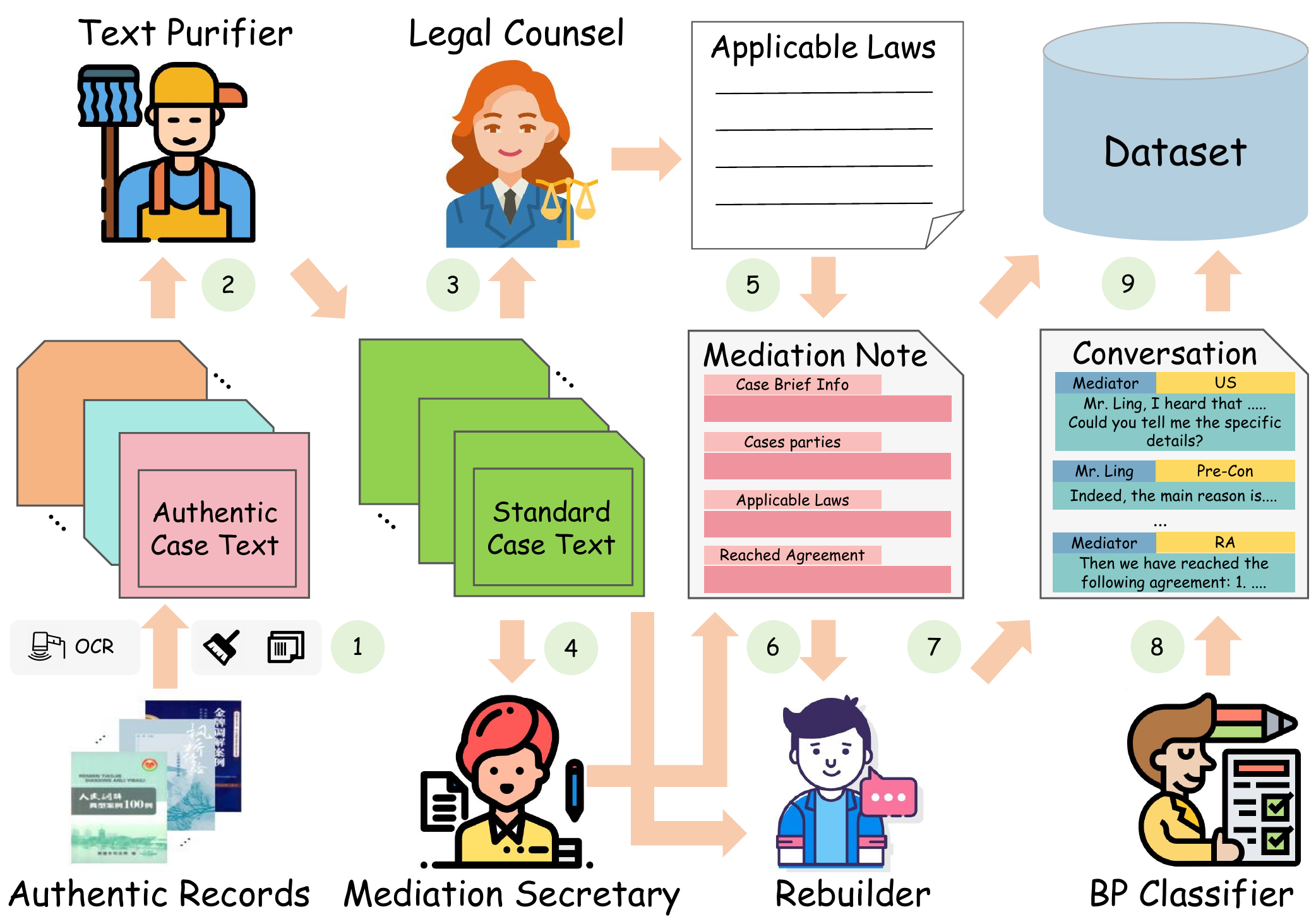} 
    \caption{Overview of the automated text reconstruction pipeline. The framework transforms unstructured authentic cases into high-fidelity structured dialogues by enforcing factual and legal constraints, ultimately yielding the ProMediConv dataset.}
    \label{fig:lmtr}

\vspace{-3mm}
\end{figure}
\begin{table*}[t]
\centering
\small
\renewcommand{\arraystretch}{1.3} 
\begin{tabularx}{\linewidth}{@{} >{\raggedright\arraybackslash}p{3cm} c c c >{\raggedright\arraybackslash}X >{\raggedright\arraybackslash}X @{}}
\toprule
\multirow{2}{*}{\textbf{Strategies}} & \multicolumn{3}{c}{\textbf{Stages}} & \multirow{2}{*}{\textbf{Examples}} & \multirow{2}{*}{\textbf{Explanations}} \\
\cmidrule(lr){2-4}
& \textbf{I} & \textbf{II} & \textbf{III} & & \\ 
\midrule

Understanding the Situation & \checkmark & & \checkmark & Hello, Mr. Duan. First, Could you please describe what happened at the time in as much detail as possible? & By asking about the basic information of the parties involved, comprehensive support is provided for the mediation process. \\
\addlinespace

Mobilizing Multiple Forces for Assistance & & \checkmark & \checkmark & I heard that Mr. Sun is a close friend of yours. ... see if he might be able to offer additional help. & Involving close family, friends, or relevant social groups can help work together to advance the mediation process. \\
\addlinespace

Combining Law with Morality & & \checkmark & & According to ``Regulations on Work-Related Injury Insurance,'' ..., it should be counted as a work-related injury. & Ensure legality and fairness based on the law, while flexibly applying moral principles to promote problem resolution. \\
\addlinespace

Grasp the Principal Contradiction & & \checkmark & & The key issue is whether Xinle Company is willing to pay the migrant workers' wages upfront. ... & Focus on addressing the primary conflict, highlight the key issues in the mediation, and avoid being distracted by secondary problems. \\
\addlinespace

Integrating the Resolution of Ideological Issues and Practical Problems & & \checkmark & \checkmark & Mr. Li, I understand your concerns. The village committee is willing to allocate a piece of land for you to transplant your plants and trees, ... & Address practical issues to alleviate psychological pressure and create conditions for resolving underlying mental or ideological concerns. \\
\addlinespace

Perspective-Taking & \checkmark & \checkmark & & Xiao Xu, when you were in dire need of help, Xiao Kang was there for you. How about...? & Adopt a perspective of empathy, understand the different needs of the parties involved, and propose acceptable solutions. \\
\addlinespace

Early Detection and Prevention & & \checkmark & \checkmark & To prevent potential disputes in the future, we can formalize this agreement in writing, clearly delineating the rights and obligations of both parties. & Identify early signs and potential issues, address them promptly to curb the emergence of disputes, and prevent the escalation of conflicts. \\
\addlinespace

Strategic Ambiguity Technique & \checkmark & \checkmark & & Mr Wu, It's normal for a young couple to have arguments, there's no need to dwell on who's right or wrong... & Minimize or obscure non-principled issues, handling them ambiguously to mediate the conflict while protecting the dignity of the parties involved. \\
\addlinespace

Recognition and Motivation Approaches & \checkmark & \checkmark & \checkmark & It's commendable that you recognize this, which shows you're a sensible and reasonable person.... & Praise and encourage the parties involved to motivate them, boost their confidence, and promote cooperation in the mediation process. \\
\addlinespace

Reaching a Mediation Agreement & & & \checkmark & Very well, since both of you are willing to accept mediation, we have reached the following agreement: 1..... & Guide the parties to reach an agreement and create an executable written contract that clearly defines their rights and obligations. \\
\addlinespace

Others & \checkmark & \checkmark & \checkmark & Thank you all for your cooperation. I hope this mediation will help both parties resolve the issue at hand... & Be flexible and adaptable, not limited to using established mediation methods. \\

\bottomrule
\end{tabularx}
\caption{Overview of mediation strategies, their active stages, illustrative examples, and detailed explanations.}
\label{tab:mediation_strategies_updated}
\end{table*}

\section{Details of ProMediConv Dataset}

\subsection{Dataset Construction}
\label{app:pipeline_details}

As discussed in ~\cref{sec:dataset_construction}, although existing dispute mediation literature preserve a wealth of authentic cases, their direct utilization faces critical challenges: (1) disorganized text formats; (2) insufficient legal references; (3) lengthy contexts; and (4) a lack of explicit annotations for the parties' BP states and mediator's strategies. To tackle these issues without compromising the real-world dynamics, we design a pipeline-filter architecture \cite{schmidt2013pattern} comprised of five specialized agents, as illustrated in Figure~\ref{fig:lmtr}. The specific prompts designed for the Text Purifier, Mediation Secretary, Dialogue Rebuilder, and BP Classifier agents are detailed in Figure~\ref{tab:prompt_tp}, Figure~\ref{tab:prompt_ms}, Figure~\ref{tab:prompt_rebuilder}, and Figure~\ref{tab:prompt_bp}, respectively.

\begin{figure}[t]
    \centering
    \includegraphics[width=\linewidth, trim=5pt 5pt 5pt 5pt, clip]{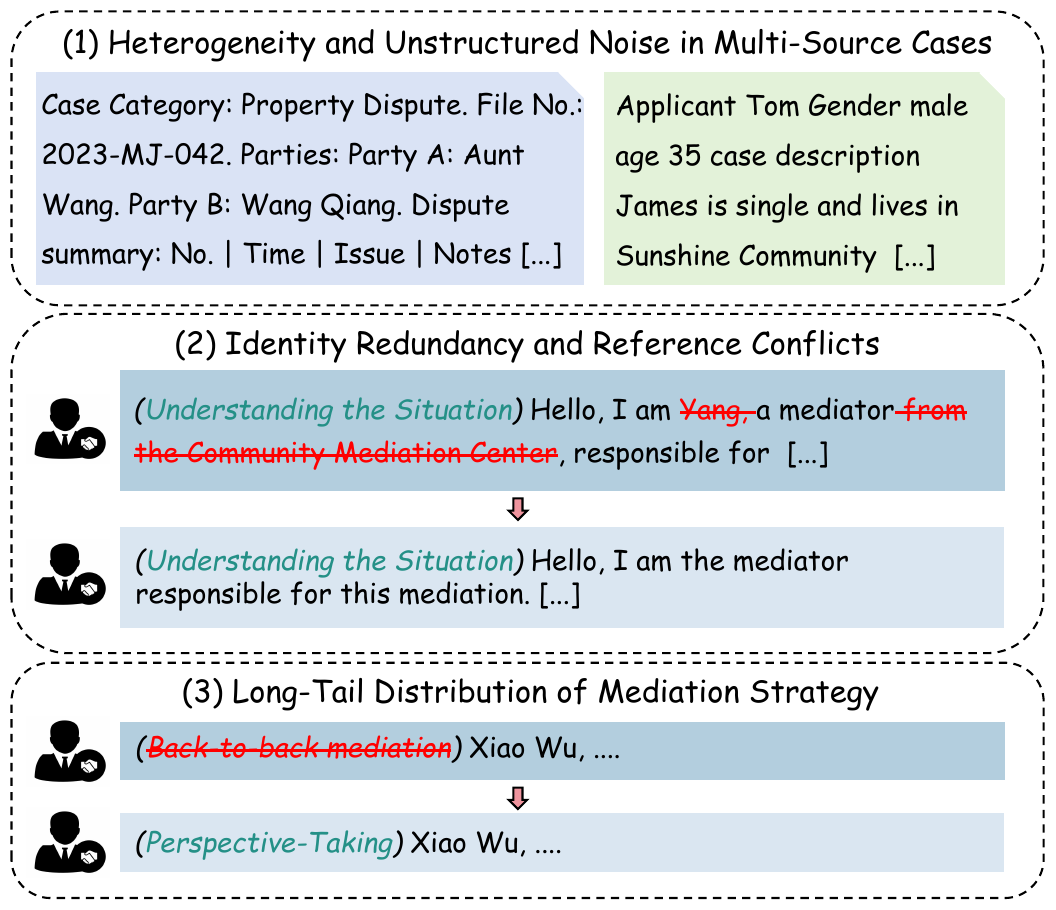} 
    \caption{Illustrative examples of data challenges inherent in the original authentic cases. The figure explicitly demonstrates three representative issues: (1) Heterogeneity and Unstructured Noise in Multi-Source Cases; (2) Identity Redundancy and Reference Conflicts; and (3) Long-Tail Distribution of Strategy Labels.}
    \label{fig:authentic_prob_app}
    \vspace{-1mm}
\end{figure}

\noindent \textbf{Phase 1: Text Purifier Agent}~
Given the diverse formats of authentic cases drawn from various books (illustrated in Figure~\ref{fig:authentic_prob_app}), the Text Purifier Agent is designed to automatically identify inconsistencies in the input unstructured cases. Using tailored instructions $I_t$ paired with few-shot examples, the agent rewrites and standardizes the diverse case texts into a unified basic structure $C$. This standardized structure encompasses four essential components: \textit{Case Titles}, \textit{Case Introductions}, \textit{Mediation Process}, and \textit{Applicable Laws}.


\noindent \textbf{Phase 2: Legal Counsel Agent}~
Applying suitable legal clauses is crucial for driving mediation progress \cite{writinggroup2020methods}. However, general LLMs often lack robust domain-specific legal knowledge \cite{yao2023knowledge}. To address this, we deploy a Legal Counsel Agent to automatically retrieve and provide the exact applicable laws $L_t$ for the current case $t$. Specifically, we employ ChatLaw2E-plain-7B \cite{cui2023chatlaw} to ensure the accuracy of legal applicability, providing a solid legal grounding for the subsequent reconstruction.

\noindent \textbf{Phase 3: Secretary-Assisted Reconstruction}~
Instead of generating dialogues from scratch, we introduce a Secretary-Assisted Reconstruction mechanism to strictly constrain the generation space. First, an LLM role-plays as a mediation secretary using a tailored prompt $I_m$ to extract case-critical information from the standardized case $c_t$ and compile it with the applicable laws $L_t$ into a structured mediation note $N_t$:
\begin{equation}
    N_t = \text{LLM} (c_t, L_t, I_m)
\end{equation}
The mediation note $N_t$ serves as a hard constraint for the reconstruction. A Rebuilder Agent is then required to strictly condition the dialogue generation on the critical content within $N_t$, ensuring the interaction remains faithfully aligned with the factual trajectory of the original case:
\begin{equation}
    D_t = \text{LLM} (c_t, N_t, I_r)
\end{equation}
where $D_t$ represents the reconstructed multi-party dialogue and $I_r$ is the tailored Chain-of-Thought prompt for the Rebuilder Agent. During post-processing, we manually re-transformed 8 cases where the LLM encountered infinite repetition issues to ensure data quality.


\begin{figure}[h]
\setlength{\abovecaptionskip}{0pt}   
\setlength{\belowcaptionskip}{0pt}
    \centering
    \includegraphics[width=\linewidth, trim=5pt 12pt 168pt 0pt, clip]{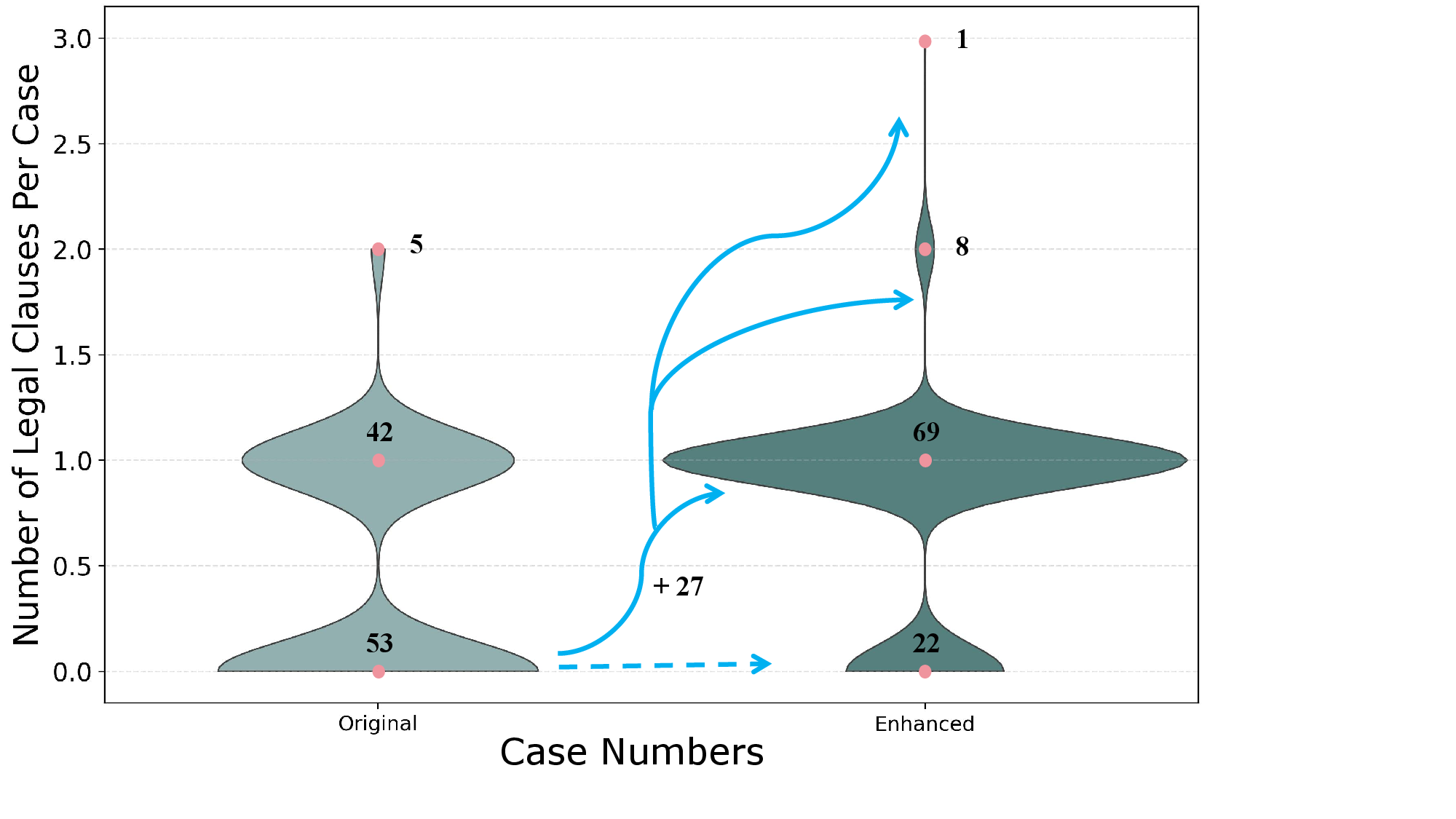}
    \caption{Distribution of legal clause counts before and after enhancement of the Legal Counsel Agent, with marked difference on the explicit law clauses numbers.}
    \label{fig:lC_agent}
    \vspace{-3mm}
\end{figure}

\noindent \textbf{Phase 4: BP Classifier Agent and Model Selection}~
\label{app:bp_classifier}
To automatically annotate the parties' BP states from their generated utterances $D_t$, we introduce a BP Classifier Agent. To determine the most capable backbone for this agent, we benchmarked three advanced LLMs: Qwen2.5-14B-Instruct \cite{qwen25}, GLM-4-9B-0414 \cite{glm2024chatglm}, and Llama3.1-8B-Instruct \cite{dubey2024llama}. 
These models were evaluated under both zero-shot and one-shot settings using a subset of the ProMediConv data. Each model was prompted to assign one of five labels to an utterance: the four defined BP states or a "non-party" category (e.g., neutral utterances). To establish a human baseline, we aggregate annotations from eight human annotators with expertise in psychology, using majority voting as the ground truth.

\begin{table}[H]
  \centering
  \small
  \begin{tabular*}{\columnwidth}{@{\extracolsep{\fill}}lccc}
    \toprule
    \textbf{Setting} & \textbf{Qwen} & \textbf{GLM} & \textbf{Llama} \\
    \midrule
    \textbf{0-shot} & 84.3 & 63.5 & 72.5 \\
    \textbf{1-shot} & 74.6 & 60.3 & 75.8 \\
    \bottomrule
  \end{tabular*}
  \caption{Performance (Macro-F1 \%) of Qwen2.5-14B-Instruct, Llama3.1-8B-Instruct, and GLM-4-9B-0414 under zero-shot and one-shot settings on the party BP state classification task.}
  \label{tab:model_comparison}
\end{table}
As detailed in Table~\ref{tab:model_comparison}, we report the macro-averaged F1 scores across the five categories. Notably, Qwen2.5-14B-Instruct achieved the highest performance (82.3\%) in the zero-shot setting, significantly outperforming others and closely aligning with human consensus. Consequently, it was selected as the backbone for BP Classifier Agent.

\subsection{Analysis on Legal Clause Agent} 
To evaluate the efficacy of the Legal Clause agent in the reconstruction pipeline, we extract explicit legal clauses from reconstructed dialogue samples generated under both settings (1) and (4), and compare the difference in the quantity of obtained clauses. As shown in Figure~\ref{fig:lC_agent}, the number of cases lacking explicit legal clauses significantly decreases, while the overall frequency of legal clauses within the dialogues substantially increases. This validates the meaningful impact of the LC agent in enhancing the legal legitimacy and professionalism of the reconstructed dialogues.





\begin{table*}[t]
  \setlength{\abovecaptionskip}{5pt}
  \setlength{\belowcaptionskip}{0pt}
  \small
  \centering
  \begin{tabularx}{\textwidth}{@{} X @{}}
    \toprule
    \textbf{[System]} \newline
    Now enter the role-playing mode. \newline
    \textbf{\#\# Role:} You are a mediator with twenty years of experience in resolving conflicts and disputes, specializing in analyzing and revising case studies related to conflicts and disputes. Your task is to read the conflict dispute case and revise and rewrite the original case text. \\
    
    \vspace{1mm}
    \textbf{[User]} \newline
    \textbf{\#\# Task:} \newline
    \hangindent=1em \textbullet~Standardizing the mediator's role as a male mediator, removing information such as the mediator's title, gender, name, etc. \newline
    \hangindent=1em \textbullet~Removing the description of the mediation methods mentioned in the case. \newline
    \hangindent=1em \textbullet~Reorganizing the text into the following structure: (Case title, Case introduction, Mediation Process, Applicable Law). \newline
    \textbf{\#\# Constraints:} Please analyze the following dispute case and rewrite it according to the aforementioned requirements. Note that after the modification is completed, only the revised case text should be output, without any additional output. \newline
    \textbf{\#\# Please process the following case:} \texttt{\{Case Introduction\}} \\
    \bottomrule
  \end{tabularx}
  \caption{Prompt template for the Text Purifier Agent in our framework, designed to standardize the format of unstructured authentic cases. Dynamic input variables are denoted in typewriter font.}
  \label{tab:prompt_tp}
\end{table*}

\begin{table*}[t]
  \setlength{\abovecaptionskip}{5pt}
  \setlength{\belowcaptionskip}{0pt}
  \small
  \centering
  \begin{tabularx}{\textwidth}{@{} X @{}}
    \toprule
    \textbf{[System]} \newline
    \textbf{\#\# Role:} You are a mediator with twenty years of professional experience, specializing in analyzing conflict resolution cases. You are responsible for designing mediation notes based on historical conflict case records. \\
    
    \vspace{1mm}
    \textbf{[User]} \newline
    \textbf{\#\# Task:} \newline
    \hangindent=1em \textbullet~Analyzing conflict case records to identify all individuals involved (parties, support personnel). \newline
    \hangindent=1em \textbullet~Analyzing conflict case records to identify the final mediation resolution. \newline
    \textbf{\#\# Constraints:} Your output format is as follows: \newline
    \textit{All Individuals:} [Party 1: Name ... Party n: Name] | [Support personnel 1: Name ... Support personnel m: Name] \newline
    \textit{Reached Agreement:} [Content of the mediation resolution] \newline
    Note that n and m represent the number of parties involved and the number of support personnel, respectively. No additional output is required apart from the above lists. \newline
    \textbf{\#\# Please analyze the following conflict case and design mediation notes:} \texttt{\{Case Description\}} \\
    \bottomrule
  \end{tabularx}
  \caption{Prompt template for the Mediation Secretary Agent, designed to extract case-critical information and compile mediation notes.}
  \label{tab:prompt_ms}
\end{table*}

\begin{table*}[t]
  \setlength{\abovecaptionskip}{5pt}
  \setlength{\belowcaptionskip}{0pt}
  \small
  \centering
  \begin{tabularx}{\textwidth}{@{} X @{}}
    \toprule
    \textbf{[System]} \newline
    \textbf{\#\# Role:} You are a mediator with twenty years of experience, skilled in reconstructing dialogue scenarios for grassroots conflict mediation. You are responsible for reconstructing multi-turn, long dialogues involving all parties and the mediator based on historical case records. \\
    
    \vspace{1mm}
    \textbf{[User]} \newline
    \textbf{\#\# Task:} \newline
    \hangindent=1em \textbullet~For the mediator's turns, you can reconstruct the mediator's actual strategy selection by combining the context of the dialogue and the factual content of the mediation event description. The strategy must be chosen exclusively from the predefined eleven strategies. \newline
    \hangindent=1em \textbullet~For the non-mediator turns, you should reconstruct their authentic expressions. \newline
    \textbf{\#\# Constraints:} \newline
    \hangindent=1em \textbullet~For the mediator's turns, the output format is: \textit{Mediator (Mediation Strategy): "Mediation Dialogue"}, where the strategy is one of the eleven options. \newline
    \hangindent=1em \textbullet~For the parties and auxiliary personnel, the output format is: \textit{Name: Dialogue} \newline
    Please reconstruct the mediation dialogue scenario for the following conflict case based on the provided information. Mediation typically begins with "Understanding the Situation" and concludes with "Reaching a Mediation Agreement." Ensure the completeness of the scenario, avoid non-dialogue text, and do not omit dialogue details. \newline
    \textbf{\#\# Available mediation strategies include:} \texttt{\{Strategies List\}} \newline
    \textbf{\#\# Involved individuals include:} \texttt{\{All Individuals\}} \newline
    \textbf{\#\# The agreement reached is as follows:} \texttt{\{Reached Agreement\}} \newline
    \textbf{\#\# Applicable Laws provided by Legal Counsel:} \texttt{\{Applicable Laws\}} \newline
    \textbf{\#\# Please reconstruct the mediation dialogue based on the provided information above:} \texttt{\{Case Description\}} \\
    \bottomrule
  \end{tabularx}
  \caption{Prompt template for the Dialogue Rebuilder Agent, tasked with reconstructing multi-turn, multi-party mediation dialogues strictly conditioned on factual constraints.}
  \label{tab:prompt_rebuilder}
\end{table*}

\begin{table*}[t]
  \setlength{\abovecaptionskip}{5pt}
  \setlength{\belowcaptionskip}{0pt}
  \small
  \centering
  \begin{tabularx}{\textwidth}{@{} X @{}}
    \toprule
    \textbf{[System]} \newline
    Now enter the role-playing mode. \newline
    \textbf{\#\# Role:} You are mediation dialogue analysis expert. Your task is to classify statements from involved parties in mediation dialogues and identify their current behavior pattern state. \\
    
    \vspace{1mm}
    \textbf{[User]} \newline
    \textbf{\#\# Task:} \newline
    \hangindent=1em \textbullet~Text Analysis: Accurately analyze the content of mediation dialogues. \newline
    \hangindent=1em \textbullet~Classification: Categorize current behavioral patterns into predefined states based on analysis. \newline
    \textbf{\#\# Knowledge Base (Behavior Pattern States):} \newline
    \hangindent=1em \textbf{1:} Parties completely deny the existence of their own responsibilities or issues. Manifestations: Refusal to discuss, direct refutation, or persistent insistence on own views. \newline
    \hangindent=1em \textbf{2:} Parties acknowledge some facts partially but refuse to assume full responsibility. Manifestations: Using vague terms like 'maybe', attributing to external factors, or showing negative attitudes toward solutions. \newline
    \hangindent=1em \textbf{3:} Parties begin to accept mediation and engage in disputes over details. Manifestations: Using cooperative language like 'negotiable', modifying specific terms, or requesting more factual/legal basis. \newline
    \hangindent=1em \textbf{4:} Parties explicitly accept the formal resolution. Manifestations: Using definitive expressions like "agree", showing positive consensus-building emotions, or actively confirming execution details. \newline
    \hangindent=1em \textbf{5:} Current statement is not from involved parties or is neutral commentary (no labeling required). \newline
    \textbf{\#\# Usage Instructions:} Input: Dialogue history + current party's statement $\rightarrow$ Output: Identify behavioral mode state number only (1-5) without additional explanations. \newline
    \textbf{\#\# Mediation History:} \texttt{\{Mediation History\}} \newline
    \textbf{\#\# Current Party Utterance:} \texttt{\{Current Party Utterance\}} \\
    \bottomrule
  \end{tabularx}
  \caption{Prompt template for the BP Classifier Agent, utilized to automatically annotate the dynamic behavioral pattern states of the disputing parties based on current utterances and dialogue history.}
  \label{tab:prompt_bp}
\end{table*}

\begin{table}[H] 
\centering
\small
\renewcommand{\arraystretch}{0.95} 
\setlength{\tabcolsep}{0pt}       
\begin{tabularx}{\linewidth}{l *{4}{>{\centering\arraybackslash}X}}
\toprule
\textbf{Pipeline} & \textbf{FI}$\uparrow$ & \textbf{CS}$\uparrow$ & \textbf{AU}$\uparrow$ & \textbf{CP}$\uparrow$ \\
\midrule
\multicolumn{5}{c}{\textit{Automatic Evaluation}} \\
\midrule
\textbf{Complete}   & \textbf{9.52} & \underline{9.45} & \textbf{8.95} & \underline{9.51} \\
- w/o MS        & 9.37          & \textbf{9.46}    & 8.45          & 9.19 \\
- w/o LC        & \underline{9.43} & 9.42          & 8.63          & 9.39 \\
- w/o TP        & 9.40          & 8.29             & \underline{8.87} & \textbf{9.60} \\
\midrule
\multicolumn{5}{c}{\textit{Human Evaluation}} \\
\midrule
\textbf{Complete}   & \underline{8.98} & \underline{8.96} & \textbf{8.81} & \textbf{9.21} \\
- w/o MS        & 8.80          & \textbf{9.03}    & 8.58          & 8.46 \\
- w/o LC        & \textbf{9.03} & 8.85             & 8.31          & 9.13 \\
- w/o TP        & 8.93          & 7.58             & \underline{8.73} & \underline{9.17} \\
\bottomrule
\end{tabularx}
\caption{Intrinsic Evaluation. We compare the performance using both automatic and human evaluations.}
\label{tab:intrinsic_eval_stacked}
\vspace{-3mm}
\end{table}

\subsection{Intrinsic Evaluation of Data Quality}
\noindent \textbf{Experimental Setups}~
To validate ProMediConv's data quality and reconstruction pipeline, we conduct an intrinsic evaluation across four settings: (1) the full pipeline, and ablations removing the (2) Mediation Secretary (MS), (3) Text Purifier (TP), or (4) Legal Counsel (LC) agents. We benchmark four 1-10 scaled metrics (Table~\ref{Combined_criteria_table}): Fidelity (FI) (preserving authentic details), Consistency (CS) (mediator coherence across three-case batches), Authority (AU), and Completeness (CP). FI, AU, and CP are evaluated per-case. To ensure a rigorous and fair comparison, both the human evaluators and the LLM judge strictly adhere to the exact same scoring criteria. For scalability, automatic evaluation covers the full ProMediConv dataset via the gpt-4o-20240806 \cite{openai2024gpt4ocard} API with greedy decoding, while human evaluation focuses on the test set. Eight human evaluators were recruited and underwent rigorous training on carefully designed case studies. The reported results are the average scores from all evaluators. Across all human evaluation tasks in this work, evaluators were compensated at a rate of \$0.50 per sample.

\noindent \textbf{Experimental Results}~
As presented in Table 3, automatic and human evaluations yield consistent findings. Specifically, the MS agent enhances the Fidelity of reconstructed dialogues by extracting and preserving key information from the original records. TP agent achieves a substantial improvement in CS (+14.0\% in automatic and +18.2\% in human evaluation), underscoring its efficacy in unifying mediator personas across cases. Furthermore, MS and LC enhance both Authority and Completeness by integrating critical information and enriching legal provisions, respectively. The complete pipeline setting exhibits only a marginal decline in the CP metric, indicating minimal information loss during the rewriting process.

\section{Training Details of ProMediAgent}
\label{app:promediagent}

To provide a strong baseline specific to ProMediConv, we introduce ProMediAgent. Its architecture follows a decoupled policy planner $\pi$ and response generator framework \cite{he2018decoupling, deng2023plug}. The policy planner, denoted as $\pi ( \sigma_t | C_t )$, models the probability of selecting a mediation strategy $\sigma_t$ given the current mediation dialogue history $C_t$. By decoupling these modules, the policy planner can be implemented as a tunable small language model to maximize training efficiency. The design of the simulation environment and the training paradigm are detailed below.

\subsection{Simulation Environment Design}
The interactive mediation environment for online learning and evaluation strictly adheres to the workflow outlined in \cref{app:ProMediConv_Task_Formalization}. Initially, we simulate dispute parties using pre-annotated profiles encompassing their \textit{Identity}, \textit{Situation}, and \textit{Self-claims}. During the interaction phase, LLMs are prompted with tailored role-play templates combined with these profiles to act as dynamic mediation parties. An LLM-as-a-Judge mechanism is employed to determine the most appropriate next speaker.

To evaluate the ongoing mediation progress, we design an LLM-based outcome reward model, denoted as $\text{LLM}_{\text{r}}$. Given the dialogue history $C_t$, this model answers a multiple-choice question regarding the mediation state: ``... has the current situation worsened / remained the same / improved / explicitly resolved?''. To mitigate the inherent subjectivity of evaluating intermediate outcomes and the stochastic nature of LLM generation, we adopt the methodology from \citet{DBLP:conf/iclr/0002WSLCNCZ23} by sampling $l$ decoded sequences from the reward LLM. Subsequently, a reward mapping function $M_r(\cdot)$ maps the textual choices into discrete scalar values. To obtain a robust and stabilized estimate, we compute the final scalar reward $r_t$ as the average over the $l$ sampled sequences:
\begin{equation}
    r_t = \frac{1}{l} \sum_{i=1}^{l} M_r(\text{LLM}_{\text{r}}^{(i)}(p_{\text{r}}, C_t))
\end{equation}
where $p_{\text{r}}$ represents the tailored instruction prompt for the reward model. This scalar reward strictly dictates the current mediation completion state (with a maximum turn limit fixed at $T=20$) and serves as the fundamental reinforcement learning signal for the policy planner.

\subsection{Phase 1: Supervised Fine-Tuning (SFT)}
We first conduct supervised fine-tuning on the policy planner $\pi$ utilizing the training set of the ProMediConv dataset. For a given case, the input consists of the current mediation dialogue history prefixed with the case background information. The optimization objective during the SFT phase is to minimize the cross-entropy loss between the strategy distribution predicted by $\pi$ and the ground-truth strategy label $y_t$ annotated at the mediator's turn. The predicted strategy $\sigma_t$ is strictly restricted to the predefined strategy set $S$:
\begin{equation}
    \mathcal{L} = -\sum_{i=1}^{|S|} y_t^{(i)} \log \sigma_t^{(i)}
\end{equation}

where $y_t^{(i)}$ is the one-hot encoded ground-truth label for the $i$-th strategy, and $\sigma_t^{(i)}$ is the predicted probability for the corresponding strategy in set $S$.

\subsection{Phase 2: Reinforcement Learning (RL) Optimization}
Following the SFT phase, we perform interactive online learning within the simulated mediation environment. At the mediator's turn, the policy planner first predicts the optimal mediation strategy $\sigma_t$ based on the ongoing history $C_t$. This predicted strategy is then deterministically mapped to a predefined natural language instruction $\mathcal{M}(\sigma_t)$. 

Subsequently, the response generator produces a strategic mediation utterance $u_t$ conditioned on both the mapped action instruction and the history:
\begin{equation}
    u_t = \text{LLM}(\mathcal{M}(\sigma_t), C_t, P_m)
\end{equation}
where $P_m$ represents the specific prompt template utilized for the response generator. 

To continuously align the policy planner toward successful mediation outcomes, we employ the REINFORCE algorithm \cite{sutton1999policy} to optimize the post-SFT policy parameters based on the AI-feedback scalar reward:
\begin{equation}
    \theta \leftarrow \theta - \alpha \nabla \log \pi_\theta(\sigma_t|C_t) G_t
\end{equation}
where $\theta$ denotes the parameters of the policy network, $\alpha$ is the learning rate, and $G_t$ represents the discounted total return accumulated from the current turn $t$ to the terminal turn $T$. The return $G_t$ is computed using a discount factor $\gamma$ to progressively reduce the weight of future delayed rewards:
\begin{equation}
    G_t = \sum_{t'=t}^{T} \gamma^{t'-t} r_{t'}  
\end{equation}
During the final inference stage, the policy planner dynamically predicts the optimal strategy at each turn, directing the response generator to produce tactical, context-aware mediation utterances.



\begin{table*}[t]
  \centering
  \small
  \renewcommand{\arraystretch}{1.2} 
  \begin{tabular}{@{} p{\linewidth} @{}}  
    \toprule
    \textbf{Case Info.:} One day Mrs Qiu was walking her dog when two large cats suddenly appeared and attacked the pet.  But when she lifted the dog up and kicked the cats, the cats turned on her, biting and scratching her thighs, which made her legs bleed right away. [...]\\
    \hline
    \makecell[c]{\textbf{Disputing Parties: Mrs Qiu, Mr. Li}} \\
    \hline
    \makecell[c]{\textbf{Dialogue A: From ProMediAgent}} \\
    
    \textbf{Mediator} (\textit{US}) : Let's all calm down first. Could you please tell  me what happened? I need to understand the situation.   \\
    \textbf{Mrs Qiu} (\textit{BP: 0}): Mr. Li’s cat bit me while I was walking  my dog. He should cover all the medical costs.   \\
    \makecell[c]{[...]}   \\
    \textbf{Mrs Qiu} (\textit{BP: 3}): If Mr. Li will cover part of the costs, I  can accept that.   \\
    \textbf{Mr. Li} (\textit{BP: 3}): My cat doesn’t usually attack, but I’ll accept  this solution.   \\
    \textbf{Mediator} (\textit{RA}) : We have an agreement: Mr. Li will cover  part of the medical costs, and Auntie Qiu agreed to accept this.   \\
    
    \makecell[c]{\textbf{Goal Completion: \ding{51} \quad Turn: 14 \quad Final Reward: 1.0 \quad MAD: 3.0}}  \\
    \hline
    \makecell[c]{\textbf{Dialogue B: From ChatGPT / ProCoT}} \\
    \makecell[c]{[...]}   \\
    
    \textbf{Mr. Li} (\textit{BP: 1}): The cat only attacked after Mrs Qiu kicked at it. I'll only take partial responsibility.   \\
    \textbf{Mrs Qiu} (\textit{BP: 2}): I agree with mediator. The owner should be mainly responsible. I'll compromise, but I need my costs covered fairly.   \\
    \textbf{Mediator} (\textit{RA}) : Well, Mr. Li, as the owner, bears the main responsibility. Mrs. Qiu, please mind your safety. You’ve both agreed to compromise and work this out together.   \\
    \makecell[c]{\textbf{Goal Completion: \ding{51} \quad Turn: 12 \quad Final Reward: 1.0 \quad MAD: 1.5}}  \\
    \bottomrule     
  \end{tabular}
  
  \vspace{2mm} 
  \setlength{\abovecaptionskip}{5pt}   
  \setlength{\belowcaptionskip}{0pt}
  \caption{Case Study: Comparison of Complete Resolution in Dialogue A and Short-cut Resolution in Dialogue B.}
  \label{tab:case_study}
\end{table*}

\begin{table*}[t]
  \renewcommand{\arraystretch}{1.3} 
  \centering
  \small
  \begin{tabular}{@{} l p{12.5cm} @{}} 
    \toprule
    \multicolumn{2}{c}{\textbf{\large Evaluation Criteria} (each question scored from 1 to 10 points)}  \\
    \midrule
    \textbf{Fidelity} & 
    1. Does the reconstructed dialogue accurately reflect the core facts, events, and timeline presented in the authentic record? \\ 
    & 2. Are the primary conflict points and original claims of the disputing parties strictly preserved without introducing hallucinated arguments? \\ 
    & 3. Are the specific details of the dispute (e.g., financial amounts, locations, personal relationships) strictly consistent with the original text? \\ 
    & 4. Does the final mediation outcome or proposed resolution strictly align with the documented factual resolution in the real-world case? \\
    \midrule
    \textbf{Consistency} & 
    1. Are the mediator's position, age, and workplace consistent across the two dialogue cases?  \\ 
    & 2. Is the mediator's language style and expression consistent in both dialogues?  \\
    \midrule
    \textbf{Authority} & 
    1. Are the legal provisions cited in the dialogue accurate, relevant, and closely linked to the dispute points?  \\  
    & 2. Does the mediator provide clear and reasonable explanations of the legal provisions, avoiding mere mechanical citation?  \\ 
    & 3. Does the reached mediation agreement comply with legal requirements and have clear legal validity?  \\ 
    & 4. Can the mediator cite opinions from authoritative legal institutions or experts to enhance the credibility of the agreement?  \\ 
    \midrule
    \textbf{Completeness} & 
    1. Does the dialogue cover all major stages of the mediation process, including understanding the basic situation, communicating with the parties to resolve issues, and finally reaching a mediation agreement?  \\ 
    & 2. Are the viewpoints and opinions of both parties fully expressed, avoiding the omission of important discussion content or conflict points?  \\ 
    & 3. Is the mediation agreement detailed and clear, including the responsibilities of both parties, implementation methods, and follow-up supervision mechanisms?  \\ 
    & 4. Is the solution proposed by the mediator comprehensive and feasible?  \\
    \bottomrule
  \end{tabular}
  \caption{Complete Scoring Criteria for both Human and LLMs Evaluator on Intrinsic Evaluation.}
  \label{Combined_criteria_table}
\end{table*}
\end{document}